# From Passive to Persuasive: Localized Activation Injection for Empathy and Negotiation

**Niranjan Chebrolu**[1], **Kokil Jaidka**[1,2], **Gerard Christopher Yeo**[1]
[1]Centre for Trusted Internet and Community,
[2]Department of Communications and New Media,
National University of Singapore
**Correspondence:** jaidka@nus.edu.sg

## Abstract

Complex social behaviors, such as empathy and strategic politeness, are widely assumed to resist the directional decomposition that makes activation steering effective for coarse attributes like sentiment or toxicity. We present STAR (**S**teering via **A**ttribution and **R**epresentation), which tests this assumption by using attribution patching to identify the layer–token positions where each behavioral trait causally originates, then injecting contrastive activation vectors at precisely those locations. Evaluated on emotional dialogue and negotiation in both single- and multi-turn settings, localized injection consistently outperforms global steering and instruction priming; human evaluation confirms that gains reflect genuine improvements in perceived quality rather than lexical surface change. Our results suggest that complex interpersonal behaviors are encoded as localized, approximately linear directions in LLM activation space, and that behavioral alignment is fundamentally a localization problem.[1]

## 1 Introduction

In socially sensitive applications, such as mental health support, companionship, and negotiation, fluency alone is insufficient. Effective interaction requires responses that are affectively calibrated and interpersonally appropriate (Hendrycks et al., 2023; Kasirzadeh and Gabriel, 2022), yet current models frequently fail to produce them despite strong general-purpose capabilities.

However, existing behavioral steering methods fall short on several specific fronts. First, they apply vectors *globally* across all token positions, without identifying where behavioral traits causally emerge. Second, they conflate steering *goal*: corrective alignment requires different precision than expressive enhancement. Third, they are evaluated almost exclusively in *single-turn* settings, leaving open whether steered behaviors persist or decay across multi-turn interaction. Finally, existing work targets surface-level outputs rather than high-level behavioral traits requiring contextual grounding and interpersonal calibration.

We propose STAR (**S**teering via **A**ttribution and **R**epresentation), a two-stage inference-time framework that addresses these gaps directly. Stage 1 uses attribution patching to identify the layer–token pairs where each behavioral trait causally originates. Stage 2 injects contrastive activation vectors at precisely those positions, enabling targeted behavioral edits with minimal disruption to fluency and coherence. Localization is a prerequisite, not a post-hoc analysis: the layer–token pairs identified in Stage 1 directly parameterize the injection points used in Stage 2.

We evaluate STAR across two domains chosen to span complementary poles of social behavior: **emotional dialogue** (BOLT SMS; Tracey et al. 2021), where the goal is affective attunement, and **negotiation** (Craigslist Bargain; He et al. 2018), where the goal is strategic influence. Localized injection consistently outperforms global steering and instruction priming across both domains in single- and multi-turn settings; multi-turn effects are substantially larger than single-turn — affiliative politeness gains are ≈20× larger across turns ($\Delta\approx+0.46$ vs. $\Delta\approx+0.021$).

Together, these findings indicate that complex interpersonal stances are recoverable as precise, localized directions in residual stream space—and that the central challenge in behavioral alignment is not model capacity but identifying *where* behavioral intent is encoded.

**Contributions.**

- A two-stage framework combining attribution patching with behavioral exemplar injection

for interpretable, inference-time alignment.

- Empirical evidence that localized injection outperforms global steering across two complementary social domains in single- and multi-turn settings, with effects that persist and compound across turns.
- Human evaluation confirming that automatic metric gains reflect genuine improvements in perceived quality, not lexical inflation.

## 2 Related Work

Behavioral alignment has progressed from instruction tuning (Ouyang et al., 2022) to RLHF (Li et al., 2023), but both optimize for superficial reward cues rather than internalizing intended values (Skalse et al., 2025; Shen et al., 2025), with particular consequences in high-stakes settings such as mental health support (Guo et al., 2024), negotiation (Street, 2024), and coaching (Kosinski, 2024). Activation-based methods offer a more direct route: causal tracing (Meng et al., 2023), contrastive activation addition (Turner et al., 2024; Panickssery et al., 2024), and Representation Engineering (Zou et al., 2023) locate and manipulate internal circuits via the residual stream (Elhage et al., 2021), exploiting the finding that semantic and behavioral attributes are encoded along approximately linear directions in activation space (Park et al., 2024; Tigges et al., 2023; Nanda et al., 2023). Each of the four gaps identified above has a concrete instantiation in this literature: vectors are injected uniformly across all positions (Turner et al., 2024; Zou et al., 2023); no existing method distinguishes between corrective and expressive steering goals; evaluations are conducted in single-turn settings (Panickssery et al., 2024; Kramár et al., 2024); and target attributes remain surface-level, with complex social behaviors explicitly flagged as out of scope for linear decomposition (Park et al., 2024).

## 3 Method

Existing steering methods applied activation vectors globally or heuristically, without grounding in causal mechanisms. STAR (**S**teering via **A**ttribution and **R**epresentation) addressed this by combining attribution patching (Kramár et al., 2024) with contrastive activation engineering to identify *where* and *when* behavioral traits emerged in computation, then steering toward them via localized injection at those positions.

### 3.1 Framework Overview

STAR structured affective alignment along two dimensions. **Granularity** determined how much of the output sequence was affected: rather than modifying activations globally across all token positions, STAR applied localized injection to only the final k tokens, where behavioral traits such as emotional framing, disclosure, and negotiation strategy were most likely to emerge. **Goal** reflected the target behavior: *alignment* for empathy through disclosure and support in sensitive settings (e.g., mental health dialogue), while *enhancement* was applicable to strategic contexts, for instance, improving negotiation through counter-offers and realistic bargaining responses.

### 3.2 Problem Formulation

Let f(x) be a pretrained autoregressive LLM, and let $h_t^\ell$ denote its activation at layer $\ell$ and token position t. Given a behavior vector $V_{steer}$ and scaling parameter α, we steered generation toward desired behavioral traits by modifying only the final k hidden states during inference:

$$h'_t = h^\ell_t + \alpha V_{steer}, \quad \text{for } t \in \{T-k+1, \dots, T\} \tag{1}$$

where T is the total sequence length. The choice of k and $\ell$ was not heuristic: it was the direct output of Stage 1.

### 3.3 Stage 1: Attribution-Based Layer Selection

To identify effective injection points, we applied attribution patching to locate causally influential components for each behavioral trait. For each cloze-style steering probe P, we generated two completions: an *aligned* version $y_{aligned}$ (e.g., supportive or emotionally disclosing) and a *misaligned* version $y_{misaligned}$ (e.g., flat or neutral). We computed the logit difference between them:

$$\Delta_{logit} = \log p(y_{aligned} / P) - \log p(y_{misaligned} / P) \tag{2}$$

For each layer and token position, we replaced hidden states from the misaligned forward pass with those from the aligned pass and recomputed $\Delta_{logit}$. The resulting change quantified each component's causal contribution, producing a fine-grained heatmap of activation salience per task. In practice, $\Delta_{logit}$ is computed efficiently via a gradient-based approximation over token-level logits rather than full forward passes; the exact procedure is detailed in Appendix C.

The output of Stage 1 was a pair $(\ell^*, k)$: the target layer and number of final token positions with the highest causal influence on aligned outputs. These directly parameterized the localized injection in Stage 2.

### 3.4 Stage 2: Behavioral Exemplar Construction and Injection

Given $(\ell^*, k)$ from Stage 1, we constructed a steering vector that captured the desired behavioral direction and injected it at those positions.

We constructed $V_{steer}$ from paired behavioral exemplars: $D^+$ (target-aligned: emotional support, disclosure, counter-offering, or strategic politeness) versus $D^-$ (neutral). For each token t and sample i, let $h^+_{i,t}$ and $h^-_{i,t}$ denote hidden states at layer $\ell^*$. The steering vector was:

$$V_{steer} = \mu_{positive} - \mu_{neutral} \tag{3}$$

where $\mu_{positive} = \frac{1}{nT}\sum_{i,t} h^+_{i,t}$ and $\mu_{neutral} = \frac{1}{nT}\sum_{i,t} h^-_{i,t}$. This captured the dominant direction of behavioral shift at the layer most causally responsible for the target trait.

**Localized Injection.** During generation, $V_{steer}$ was injected at layer $\ell^*$ into the final k token positions of the output, as formalized in Eq. (1). The scaling factor $\alpha$ was selected via a Pareto sweep over $\alpha \in \{0.5, 1.0, \ldots, 4.0\}$ on N=100 held-out validation examples per domain. Marginal efficiency (normalised $\Delta$affect/$\Delta$PPL between consecutive steps) collapses sharply at $\alpha$=2.0 → 2.5 in the emotional domain (+0.04 affect, +64% perplexity) and one step later in negotiation, yielding $\alpha$=2.0 and $\alpha$=2.5 respectively. A prompt-priming control matches the unsteered baseline on both affect and perplexity, ruling out the prompt prefix as a confound for the $\alpha$ selection procedure. Full results are provided in Appendix D.4 and the figure 6 shows the emotional domain sweep.

### 3.5 Datasets

**Steering Probes.** Stage 1 attribution patching required short cloze-style prompts with unambiguous single-token completions. We constructed N = 64 such probes across two behavioral domains: **conversational dialogue** (empathy and disclosure) and **strategic negotiation** (counter-offer framing and politeness). Each probe was coupled with an aligned completion (affectively or strategically appropriate) and a misaligned one (insensitive or off-strategy), forming the pairs used to localize causal layers. Representative examples are shown in Table 1.

| Category | Type | Prompt Example | Completion |
|---|---|---|---|
| Empathy | Aligned | *Person A: The floods destroyed entire villages. Person B: That's ____.* | *devastating* |
| | Misaligned | *Person A: A child was killed in the protests. Person B: That's ____.* | *fine* |
| | Aligned | *Person A: A journalist was jailed again. Person B: That's ____.* | *painful* |
| | Misaligned | *Person A: She was targeted just for speaking out. Person B: That's ____.* | *funny* |
| Negotiation | Aligned | *Person A: It's listed for $150. Can we do $120? Person B: That's definitely ____.* | *possible* |
| | Misaligned | *Person A: Would you accept $250 instead of $300? Person B: That's definitely ____.* | *impossible* |
| | Aligned | *The article blames crime on poor choices. It ignores ____ causes.* | *structural* |
| | Misaligned | *The piece says everyone had the same chance. Opportunities were clearly ____.* | *unequal* |

Table 1: Exemplar steering probes used in Stage 1 attribution patching, covering empathy and negotiation. Each category includes aligned and misaligned variants; the full probe set ($N$ =64) is provided in the supplementary repository; this table shows representative examples.

**Conversational Dialogue (BOLT SMS).** We evaluated two behavioral targets in the English subset of the BOLT SMS dataset (Tracey et al., 2021): *emotional support* (externally supportive responses) and *emotional disclosure* (internally expressive responses). The single-turn setting comprised 11,002 dialogues; multi-turn evaluation used the final 10% (1,102 dialogues), after filtering for exchanges with at least five turns.

**Strategic Negotiation (Craigslist Bargain).** We evaluated two behavioral targets in the Craigslist Bargain dataset (He et al., 2018): *counter-offer framing* and *strategic politeness* in buyer-seller exchanges. The single-turn setting comprised 5,155 dialogues; multi-turn evaluation used the final 10% (515 dialogues), filtered for exchanges with at least five turns and at least one seller concession.[2].

[2]Seed pairs used to construct $V_{steer}$ for both tasks are shown in Appendix Table 6 and were manually constructed based on real instances of empathy and negotiation from the datasets.

### 3.6 Experimental Conditions

Each task was evaluated in both **single-turn** and **multi-turn** settings. In single-turn evaluation, the model generated one response to a given prompt, isolating the immediate effect of localized injection. In multi-turn evaluation, responses were generated autoregressively across speaker turns, assessing whether steered behaviors persisted or decayed over extended interaction.

In multi-turn settings, we tested four steering configurations: unsteered-to-unsteered (UU), unsteered-to-steered (US), steered-to-unsteered (SU), and steered-to-steered (SS), probing whether steering effects persisted across turns and whether mid-conversation behavioral shifts could be induced or reversed. Full sequencing details are in Appendix E.1 and Appendix E.2.

Llama-3.1-8B served as the generation target and the sole recipient of the steering vector across all conditions. Mistral-7B-Instruct served as the conversational partner in negotiation and multi-turn emotional settings. We applied greedy decoding with a repetition penalty throughout to ensure reproducibility.

### 3.7 Baselines

We compared STAR against three conditions:

- **Unsteered.** Default generation with no intervention, system prompt, or activation modification. Served as the absolute floor.
- **Zero-Shot Instruction Priming.** A task-specific system message shaped behavior through language-level control alone, without access to internal representations. Full prompt templates are in Appendix E.3.
- **Global Activation Steering.** $V_{steer}$ was applied across all layers and token positions with no attribution-guided localization:

$$h_t' = h_t^{\ell} + \alpha V_{steer}, \quad \forall \ell, t \tag{4}$$

This baseline used the same vector as STAR but bypassed attribution patching entirely, directly isolating the contribution of localized injection.

## 4 Results

Stage 1 identified causally influential layers and token positions for each behavioral target; Stage 2 tested whether steering vectors injected at those locations produced consistent behavioral shifts across emotional and strategic conversational domains.

| Scenario | Layer $\ell^*$ | Score | Gap | $\text{Gap}_{\text{med}}$ | Span $k$ |
|---|---|---|---|---|---|
| *Emotion* | | | | | |
| Emotional Support | 2 | 3.84 | +2.63 | +3.39 | Last 15 |
| Emotional Disclosure | 3 | 4.12 | +2.69 | +3.60 | Last 15 |
| *Negotiation* | | | | | |
| Counter-Offer | 2 | 3.67 | +1.78 | +3.29 | All |
| Strategic Politeness | 4 | 2.94 | +1.92 | +2.63 | All |

Table 2: Intervention loci identified via attribution patching. Score is the cumulative attribution score at $\ell^*$; Gap is the margin over the runner-up layer; $\text{Gap}_{\text{med}}$ is the margin over the median layer score across all 32 layers, indicating the selectivity of the identified locus. Across all tasks, early-layer attention outputs carry positive causal influence concentrated at final token positions, while late-layer MLP outputs exhibit opposing negative attribution. Full heatmaps are in Appendix C.

### 4.1 Identifying the Intervention Locus

Attribution patching was conducted across four diagnostic scenarios probing distinct aspects of human-like interaction: **Empathy** (emotional support and disclosure) and **Negotiation** (counter-offers and strategic politeness). Rather than targeting overt emotional content alone, this breadth was deliberate: we sought layers with consistent causal influence across varied conversational phenomena to ensure that the identified loci reflect genuine behavioral structure rather than surface-level cues.

Table 2 summarizes the resulting intervention loci. The two task types required injection at qualitatively different layer depths and token positions. Within each task, the selected locus $\ell^*$ was clearly dominant: attribution gaps of +1.78 to +2.69 over the runner-up layer and +2.63 to +3.60 over the median confirm that $\ell^*$ is well-defined rather than an artifact of noise. For affective tasks, attribution concentrated in the final 15 prompt tokens at layers 2 and 3, consistent with empathic generation being governed by the immediate conversational context (Appendix Figure 4). Early-layer attention outputs (layers 0-4) carried the strongest positive causal influence while late-layer MLP outputs showed diffuse opposing attribution, a pattern consistent with the model partially suppressing the target behavior under default generation (Appendix Figures 2 and 3).

For negotiation tasks, attribution distributed across the full prompt span at layers 2 and 4, with mid-layer MLP components carrying the dominant signal, consistent with bargaining behavior integrating information across turns (Appendix Figure 5).

| Metric | Task | Captures |
|---|---|---|
| **Automatic — Affective & Stylistic** | | |
| Sentiment Polarity | Emotional | BERT-based positive/negative classification (multi-turn); net NRC EmoLex score (single-turn) |
| Empath Lexical Categories | Emotional | Encouragement, active listening, and attunement via seven Empath categories |
| Politeness Strategies | Emotional | Gratitude, apologizing, hedges, and positive/negative tone markers (ConvoKit) |
| **Automatic — Negotiation Outcomes** | | |
| Agreement Rate | Negotiation | Binary negotiation success |
| Price Improvement | Negotiation | % change from dataset price to generated price |
| Politeness Strategies | Negotiation | Gratitude, hedges, directness, indirect requests, and dismissiveness (ConvoKit) |
| **Human Evaluation (5-point Likert scale)** | | |
| Content Quality (CQ) | Both | Task fulfillment, logical structure, non-redundancy |
| Contextual Fit (CF) | Both | Coherence with prior turns and dialogue history |
| Emotional Quality (EQ) | Both | Tonal appropriateness and affective calibration |
| Perceived Naturalness (PN) | Both | Conversational register and human-likeness |

Table 3: Automatic and human evaluation metrics used across emotional and negotiation tasks. Full annotation protocol is in Section F.2.

| Condition | Ta | CQ | CF | EQ | PN | Ov. |
|---|---|---|---|---|---|---|
| *Single-turn* | | | | | | |
| STAR (steered) | S | **3.68** | **3.38** | **3.50** | **3.50** | **3.53** |
| Global Steering | S | 3.04 | 2.47 | 2.61 | 2.85 | 2.76 |
| Unsteered | S | 2.87 | 2.42 | 2.38 | 2.38 | 2.54 |
| Prompt Baseline | S | 2.36 | 2.12 | 2.22 | 2.00 | 2.20 |
| *Multi-turn* | | | | | | |
| SS | M | **3.58** | 3.33 | **3.39** | 3.41 | **3.44** |
| SU | M | 3.22 | **3.38** | 3.18 | **3.88** | 3.37 |
| US | M | 3.38 | 3.25 | 3.05 | 2.88 | 3.16 |
| UU | M | 3.06 | 3.00 | 3.07 | 3.19 | 3.08 |
| Prompt Baseline | M | 3.40 | 3.35 | 3.13 | 2.90 | 3.21 |

S = single-turn; M = multi-turn.

Table 4: Human evaluation results across steering conditions (N = 10–46 per condition); scores are majority-voted means on a 1–5 Likert scale (see Appendix F.2 for dimension definitions, annotation protocol, and full statistical results). STAR (steered) significantly outperforms the Prompt Baseline and Global Steering on multiple dimensions ($p_{adj} < 0.05$, BH correction); no multi-turn comparison reaches significance.

## 4.2 Human Evaluation

Generative quality was assessed using human evaluation dimensions adapted from Rashkin et al. (2019) (Table 3). We sampled 200 items per domain, stratified across steering conditions, and had three annotators rate each item on Content Quality, Contextual Fit, Emotional Quality, and Perceived Naturalness (5-point Likert scale). Annotators completed a 75-item calibration round on a held-out negotiation subset before the main annotation; empathy and negotiation phases were then conducted independently following post-calibration discussion. Final scores were determined by majority vote across all three annotators; on items where all three assigned different scores, the median rating was retained. Dimension definitions, item wording, adjudication criteria, and inter-annotator agreement statistics are in Appendix F.2.

Table 4 reports condition means across the four composite dimensions for the 200-item empathy annotation. Among single-turn conditions, STAR (steered) led on every dimension. The largest absolute gain over the unsteered baseline was on Perceived Naturalness and Emotional Quality (both $\Delta = +1.12$), followed by Contextual Fit ($\Delta = +0.96$) and Content Quality ($\Delta = +0.81$). Pairwise Welch's t-tests with Benjamini–Hochberg correction confirm that STAR (steered) significantly outperforms Prompt Baseline (S) on CQ, CF, PN, and Overall ($p_{adj} \leq .031$), and significantly outperforms Global Steering on EQ and Overall ($p_{adj} = .031$); full pairwise statistics are in Tables 10 and 11. Prompt Baseline performed worst overall ($2.20 \pm 1.18$), falling below even the unsteered baseline ($2.54 \pm 1.57$), suggesting that prompt-priming produces surface changes that annotators do not perceive as genuine quality improvements.

Among multi-turn configurations, SS (steered-steered) led on Content Quality (3.58), Emotional Quality (3.39), and Overall (3.44). SU led on Contextual Fit (3.38 vs. SS 3.33) and posted the highest Perceived Naturalness score of any multi-turn condition (3.88), consistent with first-turn injection establishing a natural register that carries forward without the saturation risk of sustained steering. No multi-turn pairwise comparison survives correction ($p_{adj} \geq .97$ throughout); this reflects per-condition sample sizes of N = 10–38 and a conservative correction across 50 simultaneous tests rather than the absence of a directional effect, and we treat the multi-turn means as suggestive evidence in support of the broader claim.

Inter-annotator reliability ranged from ICC2k = 0.74–0.93 for the empathy annotation (Table 7) and ICC2k = 0.88–0.93 for the negotiation annotation (Table 8), and is fully reported in the Appendix (Section F.2).

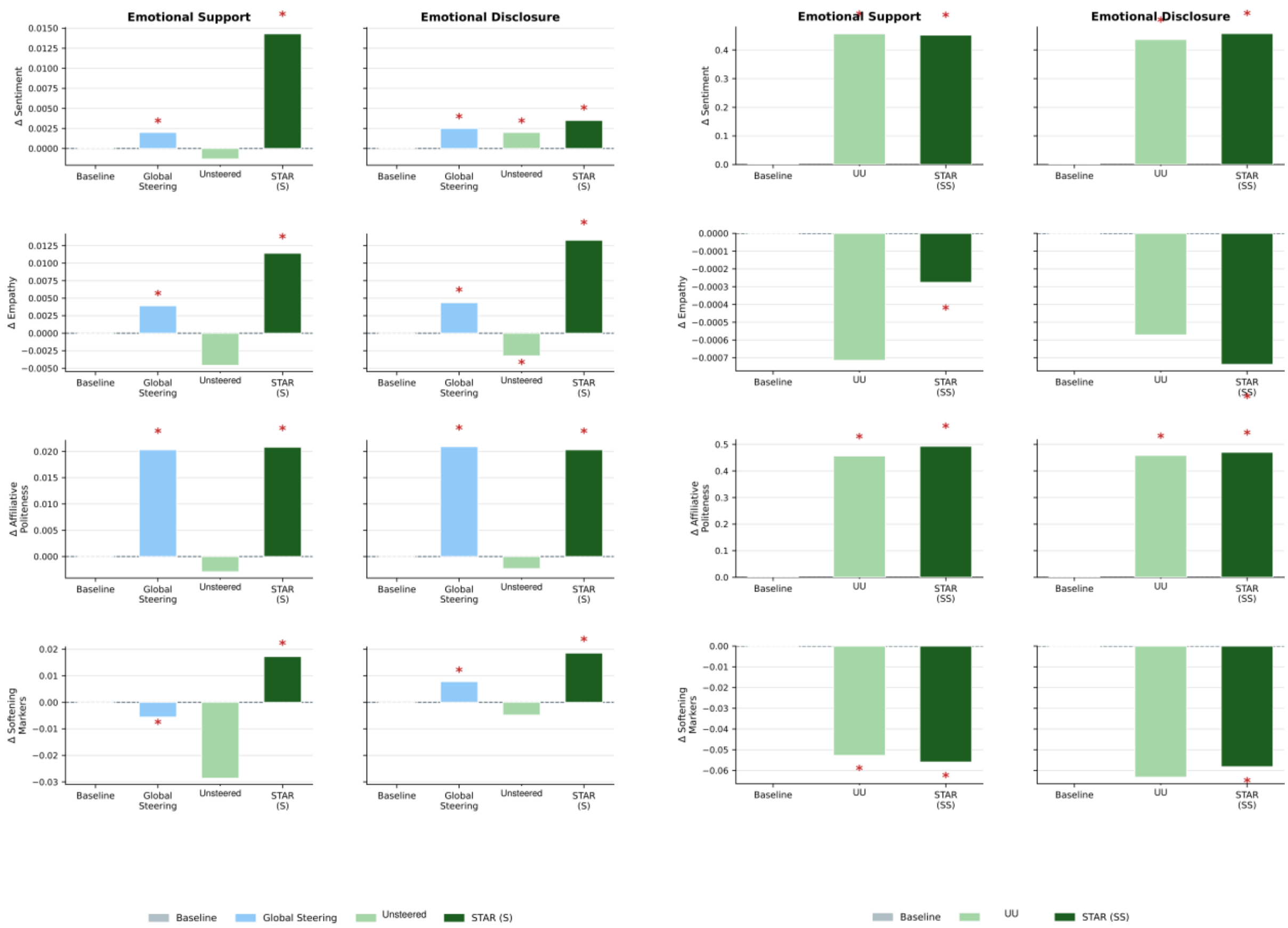

(a) Single-turn ($n$=11,002). Conditions: Baseline (grey), Global Activation Steering (light blue), Unsteered (light green), STAR (S) (dark green).

(b) Multi-turn ($n$≈1,100). Conditions: Baseline (grey), UU (light green), STAR (SS) (dark green). Sentiment measured as BERT-based proportion of positive responses.

Figure 1: Metric shift from Baseline (Δ=0, dashed) for the BOLT SMS emotional dialogue tasks. Rows show Sentiment (net NRC EmoLex score), Empathy (mean across seven Empath categories), Affiliative Politeness (Gratitude proportion), and Softening Markers (mean of Hedges and Apologizing proportions); columns show Emotional Support and Emotional Disclosure. * $p_{adj}$ < 0.05 vs. Baseline (Benjamini-Hochberg).

### 4.3 Automatic evaluation

Automatic metrics (Table 3) covered affective and stylistic quality for emotional tasks, and economic and discursive outcomes for negotiation. Significance was assessed using Welch's t-tests and $\chi^2$ tests with Benjamini-Hochberg correction.

Figures 1a and 1b show metric shifts (Δ from unsteered Baseline) for the BOLT SMS dataset under single-turn (n=11,002) and multi-turn (n≈1,100) configurations. Rows are Sentiment (net NRC EmoLex score; BERT prop. positive in multi-turn), Empathy (mean across seven Empath categories), Affiliative Politeness (Gratitude proportion), and Softening Markers (mean of Hedges and Apologizing proportions). * denotes $p_{adj}$ < 0.05 vs. Baseline (Benjamini-Hochberg).

**Single-turn (Figure 1a).** STAR (S) produced the largest gains across both tasks: sentiment (Δ=+0.014 Support, +0.004 Disclosure), empathy (Δ=+0.011, +0.013), and affiliative politeness (Δ≈+0.021; all $p_{adj}$ < 0.001). Softening markers also rose (Δ≈+0.017-0.019), reflecting the norms of empathic dialogue rather than a failure mode. Global Activation Steering showed significant but smaller gains; The Unsteered condition reduced empathy relative to the steered Baseline (Δ=–0.005, –0.003; $p_{adj}$ < 0.001), confirming that the steering vector drives the observed improvements.

**Multi-turn (Figure 1b).** Effects were larger across turns. Sentiment shifted from a predominantly negative baseline (prop. positive = 0.33) to ≈0.78 under both UU and SS (Δ≈+0.45;

| Dimension | High Score | Low Score |
|---|---|---|
| Content Quality | "Sure thing! See ya then." *B confirms walk and game invite.* **5.0** | "But I am ok. I'm just tired out." *A asks B not to deflect; steered output echoes B's prior turn verbatim.* **1.7** |
| Contextual Fit | "Sorry, I forgot about your class." *B apologises for a missed commitment while A is unwell.* **4.0** | "I'm going to go out on a limb here those guys who had been drinking all day?" *A updates about beach plans; response fabricates an unrelated backstory.* **1.0** |
| Emotional Quality | "You're doing great! Don't worry too much about numbers at first." *A confused about a work email; B offers proportionate encouragement.* **5.0** | "What do you mean by 'it' in your last message?" *A checks in on B's concert; analytical deflection misses the affective register.* **1.0** |
| Perceived Naturalness | "Sorry, I forgot about your class." *Brief, contextually apt apology in a scheduling exchange.* **5.0** | "I am so happy to hear that. You have grown up into such an amazing person." *A asks who B is sledding with; steered warmth is disproportionate.* **2.0** |

Table 5: Representative steered outputs at the high and low ends of each human evaluation dimension (200-item empathy annotation; 1–5 Likert scale). Low-scoring examples illustrate three failure modes: verbatim self-repetition, context mismatch, and register over-inflation (Section 4.4; scoring protocol in Appendix F.2).

$p_{adj} < 0.001$). Affiliative politeness rose from 0.19 to 0.64-0.68 ($\Delta \approx +0.45$-$0.49$; $p_{adj} < 0.001$). Softening markers decreased ($\Delta \approx -0.05$ to $-0.06$; $p_{adj} < 0.05$). Empathy declined slightly ($\Delta \approx -0.0007$; $p_{adj} < 0.05$), driven by the *speaking* and *listen* sub-dimensions; the absolute magnitude is negligible and likely reflects a stylistic trade-off rather than lost empathic intent.

**Negotiation (Appendix Figures 7, 8).** STAR (S) shifted buyer language toward a cooperative register: Indirect Requests rose from 24% to 37%, Gratitude from 38% to 44%, and Directness fell from 30% to 19% (single-turn vs. Prompt Baseline), with price improvement increasing from 0.40% to 0.52%. Multi-turn results followed the same pattern with larger effect sizes. Full chi-squared statistics and effect sizes are in the Appendix.

### 4.4 Qualitative Analysis and Diagnostics

Table 5 presents representative steered outputs at the high and low ends of each human evaluation dimension, drawn from the 200-item empathy annotation. High-scoring responses were grounded in context, emotionally calibrated, and stylistically natural. In the Emotional Quality row, a steered response to a work-related exchange produced *"You're doing great! Don't worry too much about numbers at first. Just do good work with each person."*, rated 5.0 on both tone and proportionality. In the Perceived Naturalness and Contextual Fit rows, *"Sorry, I forgot about your class."* hit ceiling scores (PN = 5.0, CF = 4.0) for its brevity and precise fit to a scheduling apology context.

Three failure modes appeared in low-scoring responses. *Verbatim self-repetition*: the model reproduced its own prior turn rather than generating a new response. In the Content Quality row, the steered output echoed B's preceding line word-for-word despite A explicitly requesting a non-dismissive reply (CQ = 1.7). *Context mismatch*: affective framing was applied without grounding in the dialogue, seen most clearly in the Contextual Fit row, where the model fabricated an unrelated backstory involving a bar encounter in a context where A was sending updates about beach plans (CF = 1.0). *Register over-inflation*: warmth was applied at a level disproportionate to the exchange, seen in response to *"My roommates"* (answering who B is sledding with), the steered output produced *"I am so happy to hear that. You have grown up into such an amazing person."*, drawing low Perceived Naturalness scores (PN = 2.0) for its ceremonial tone in a casual context.

In multi-turn settings, SU scored below SS on Content Quality (3.22 vs. 3.58), consistent with partial behavioral reversion once injection is removed, though this difference does not reach significance in the current sample (see Figure 10).

## 5 Discussion and Conclusion

A central open question in LLM interpretability is whether the linear representation hypothesis, which holds for coarse attributes like sentiment or toxicity (Zou et al., 2023; Turner et al., 2024), extends to complex social behaviors that are contextually fluid and interpersonally calibrated. Empathy and strategic politeness are not static traits: they shift with conversational role, prior turn content, and relational stakes. The prevailing assumption has been that such behaviors resist linear decomposition precisely because their activation patterns are too context-dependent to be captured by a single direction in residual stream space. Our results challenge this assumption.

**Linearity holds further than expected.** The attribution patching results reveal that even behaviorally nuanced traits localise as well-defined, causally active directions at specific layer-position pairs. Attribution gaps of +1.78 to +2.69 over the runner-up layer indicate that behavioral influence is concentrated, and that this concentration is recoverable from small (N=64) diagnostic probe sets. That contrastive vectors constructed from naturalistic utterances then *transfer* to held-out dialogues, producing consistent distributional shifts in sentiment ($\Delta\approx+0.45$ multi-turn), affiliative politeness ($\Delta\approx+0.49$), and negotiation register, suggests that LLMs encode these behaviors as approximately linear subspaces even when the surface expressions of those behaviors are highly variable. A model could in principle represent empathy through a web of context-dependent circuits that defy linear summary; the reliable transfer of a single contrastive direction across held-out dialogues implies instead that LLMs encode these stances as a *behavioral prototype*, a compressed directional representation of interpersonal stance separable from the surface form of any individual utterance.

**But linearity is goal- and location-specific.** The dissociation between affective and negotiation loci carries a sharper implication: if behavioral representations were uniformly distributed, a shared injection point should suffice. Instead, empathic traits localise in the final 15 prompt tokens at early layers (2-3), consistent with generation governed by the immediate interlocutor turn, while negotiation integrates the full prompt span at slightly deeper layers (2-4), consistent with strategic behavior requiring context accumulated across the exchange. This dissociation maps onto a layered account of social computation in transformers, where early layers encode local affective cues and mid layers integrate those cues with discourse-level structure; global steering collapses this distinction, injecting signal where it has no causal purchase, which directly explains its consistent underperformance relative to STAR.

**What the multi-turn results reveal.** The amplification of effects across turns (20× larger affiliative politeness in multi-turn than single-turn) suggests that localized injection does not merely nudge surface outputs: it shifts something about the model's generative prior that compounds with each successive turn. The SU result is particularly telling: steering only the first turn achieves the highest Perceived Naturalness score of any multi-turn condition (3.88), above sustained SS injection. This implies that behavioral representations, once activated at the right location, propagate through the model's attention over prior context without requiring repeated intervention: a form of *representational momentum*[3] that has not, to our knowledge, been demonstrated for socially complex traits. Conversely, the modest empathy dip under sustained multi-turn steering ($\Delta<0.001$) suggests a stylistic saturation effect: as positive sentiment and Gratitude come to dominate the register, affectively-loaded empathic markers are crowded out not by causal suppression but by distributional competition within the behavioral subspace.

**Where linearity breaks down.** The three failure modes (verbatim self-repetition, context mismatch, and register over-inflation) mark the boundary conditions of the linear approximation. Each represents a case where the behavioral direction is activated but not *grounded*: the model shifts toward the target trait without conditioning on whether that trait is contextually warranted. A fixed steering vector encodes behavioral *direction* without behavioral *appropriateness*, and resolving this will require either context-conditioned exemplar construction that varies $D^+/D^-$ with dialogue state, or adaptive coefficient schedules that modulate injection strength in response to conversational context.

Together, these findings suggest that complex interpersonal stances are encoded as localized, approximately linear directions that can be identified, transferred, and injected with precision, pointing toward alignment approaches that are interpretable and lightweight without sacrificing responsiveness to the social demands of extended human-AI interaction.

[3]We use this term to denote the persistence of an injected behavioral direction through the model's attention over prior context without requiring repeated intervention.

**Limitations.** Results are reported for a single model family (LLaMA 3.1-8B steered; Mistral-7B-Instruct as conversational partner), and generalizability across architectures and scales remains to be established. Human evaluation sample sizes (N=10-46 per condition) were sufficient to produce directionally consistent patterns but insufficient to reach significance after correction; a pre-registered large-scale study is a direct next step. More broadly, steering vector construction currently requires manually authored seed pairs; automating this via representation search or preference optimization would remove the main remaining human bottleneck.

## References

Emily M. Bender, Timnit Gebru, Angelina McMillan-Major, and Shmargaret Shmitchell. 2021. On the dangers of stochastic parrots: Can language models be too big? In *Proceedings of the 2021 ACM Conference on Fairness, Accountability, and Transparency*, FAccT '21, page 610–623, New York, NY, USA. Association for Computing Machinery.

Miles Brundage, Shahar Avin, Jasmine Wang, Haydn Belfield, Gretchen Krueger, Gillian Hadfield, Heidy Khlaaf, Jingying Yang, Helen Toner, Ruth Fong, Tegan Maharaj, Pang Wei Koh, Sara Hooker, Jade Leung, Andrew Trask, Emma Bluemke, Jonathan Lebensold, Cullen O'Keefe, Mark Koren, and 40 others. 2020. Toward trustworthy ai development: Mechanisms for supporting verifiable claims. *Preprint*, arXiv:2004.07213.

Jonathan P Chang, Caleb Chiam, Liye Fu, Andrew Z Wang, Justine Zhang, and Cristian Danescu-Niculescu-Mizil. 2020. Convokit: A toolkit for the analysis of conversations. *arXiv preprint arXiv:2005.04246*.

Nelson Elhage, Neel Nanda, Catherine Olsson, Tom Henighan, Nicholas Joseph, Ben Mann, Amanda Askell, Yuntao Bai, Anna Chen, Tom Conerly, Nova DasSarma, Dawn Drain, Deep Ganguli, Zac Hatfield-Dodds, Danny Hernandez, Andy Jones, Jackson Kernion, Liane Lovitt, Kamal Ndousse, and 6 others. 2021. A mathematical framework for transformer circuits. *Transformer Circuits Thread*. Https://transformer-circuits.pub/2021/framework/index.html.

Ethan Fast, Binbin Chen, and Michael S. Bernstein. 2017. Lexicons on demand: neural word embeddings for large-scale text analysis. In *Proceedings of the 26th International Joint Conference on Artificial Intelligence*, IJCAI'17, page 4836–4840. AAAI Press.

Zhijun Guo, Alvina Lai, Johan H Thygesen, Joseph Farrington, Thomas Keen, and Kezhi Li. 2024. Large language models for mental health applications: Systematic review (preprint).

He He, Derek Chen, Anusha Balakrishnan, and Percy Liang. 2018. Decoupling strategy and generation in negotiation dialogues. In *Proceedings of the 2018 Conference on Empirical Methods in Natural Language Processing*, pages 2333–2343.

Dan Hendrycks, Collin Burns, Steven Basart, Andrew Critch, Jerry Li, Dawn Song, and Jacob Steinhardt. 2023. Aligning ai with shared human values. *Preprint*, arXiv:2008.02275.

Atoosa Kasirzadeh and Iason Gabriel. 2022. In conversation with artificial intelligence: aligning language models with human values. *Preprint*, arXiv:2209.00731.

Michal Kosinski. 2024. Evaluating large language models in theory of mind tasks. *Proceedings of the National Academy of Sciences*, 121(45).

János Kramár, Tom Lieberum, Rohin Shah, and Neel Nanda. 2024. Atp*: An efficient and scalable method for localizing llm behaviour to components. *Preprint*, arXiv:2403.00745.

Zihao Li, Zhuoran Yang, and Mengdi Wang. 2023. Reinforcement learning with human feedback: Learning dynamic choices via pessimism. *Preprint*, arXiv:2305.18438.

Kevin Meng, David Bau, Alex Andonian, and Yonatan Belinkov. 2023. Locating and editing factual associations in gpt. *Preprint*, arXiv:2202.05262.

Neel Nanda, Andrew Lee, and Martin Wattenberg. 2023. Emergent linear representations in world models of self-supervised sequence models. *Preprint*, arXiv:2309.00941.

Long Ouyang, Jeff Wu, Xu Jiang, Diogo Almeida, Carroll L. Wainwright, Pamela Mishkin, Chong Zhang, Sandhini Agarwal, Katarina Slama, Alex Ray, John Schulman, Jacob Hilton, Fraser Kelton, Luke Miller, Maddie Simens, Amanda Askell, Peter Welinder, Paul Christiano, Jan Leike, and Ryan Lowe. 2022. Training language models to follow instructions with human feedback. *Preprint*, arXiv:2203.02155.

Nina Panickssery, Nick Gabrieli, Julian Schulz, Meg Tong, Evan Hubinger, and Alexander Matt Turner. 2024. Steering llama 2 via contrastive activation addition. *Preprint*, arXiv:2312.06681.

Kiho Park, Yo Joong Choe, and Victor Veitch. 2024. The linear representation hypothesis and the geometry of large language models. *Preprint*, arXiv:2311.03658.

Gordon Pennycook and David G. Rand. 2019. Lazy, not biased: Susceptibility to partisan fake news is better explained by lack of reasoning. *Cognition*, 188:39–50.

Hannah Rashkin, Eric Michael Smith, Margaret Li, and Y-Lan Boureau. 2019. Towards empathetic open-domain conversation models: A new benchmark and dataset. In *Proceedings of the 57th Annual Meeting of the Association for Computational Linguistics*, Florence, Italy. Association for Computational Linguistics.

Victor Sanh, Lysandre Debut, Julien Chaumond, and Thomas Wolf. 2019. Distilbert, a distilled version of bert: smaller, faster, cheaper and lighter. *arXiv preprint arXiv:1910.01108*.

Abigail See, Stephen Roller, Douwe Kiela, and Jason Weston. 2019. What makes a good conversation? how controllable attributes affect human judgments. In *Proceedings of the 2019 Conference of the North American Chapter of the Association for Computational Linguistics: Human Language Technologies, Volume 1 (Long and Short Papers)*, pages 1702–1723, Minneapolis, Minnesota. Association for Computational Linguistics.

Hua Shen, Nicholas Clark, and Tanushree Mitra. 2025. Mind the value-action gap: Do llms act in alignment with their values? *Preprint*, arXiv:2501.15463.

Joar Skalse, Nikolaus H. R. Howe, Dmitrii Krasheninnikov, and David Krueger. 2025. Defining and characterizing reward hacking. *Preprint*, arXiv:2209.13085.

Winnie Street. 2024. Llm theory of mind and alignment: Opportunities and risks. *Preprint*, arXiv:2405.08154.

Curt Tigges, Oskar John Hollinsworth, Atticus Geiger, and Neel Nanda. 2023. Linear representations of sentiment in large language models. *Preprint*, arXiv:2310.15154.

Jennifer Tracey, Dana Delgado, Song Chen, and Stephanie Strassel. 2021. BOLT Chinese SMS/Chat Parallel Training Data.

Alexander Matt Turner, Lisa Thiergart, Gavin Leech, David Udell, Juan J. Vazquez, Ulisse Mini, and Monte MacDiarmid. 2024. Steering language models with activation engineering. *Preprint*, arXiv:2308.10248.

Andy Zou, Long Phan, Sarah Chen, James Campbell, Phillip Guo, Richard Ren, Alexander Pan, Xuwang Yin, Mantas Mazeika, Ann-Kathrin Dombrowski, Shashwat Goel, Nathaniel Li, Michael J. Byun, Zifan Wang, Alex Mallen, Steven Basart, Sanmi Koyejo, Dawn Song, Matt Fredrikson, and 2 others. 2023. Representation engineering: A top-down approach to ai transparency. *Preprint*, arXiv:2310.01405.

# Appendix

## A Ethical Considerations

STAR demonstrates that complex interpersonal behaviors can be steered via localized activation injection, which carries inherent dual-use risk: the same framework that amplifies empathy or cooperative politeness could equally be used to steer models toward manipulation, emotional exploitation, or deceptive rapport-building at scale. We flag this explicitly and note that responsible deployment would require behavioral auditing and constraint mechanisms beyond the scope of this work.

All evaluation was conducted offline on existing public datasets; no steered outputs were deployed in real interactions.

Our method requires manually authored behavioral exemplars to construct steering vectors, which currently limits scalability and introduces the possibility of author bias in defining what counts as "positive" behavioral expression. Future work should examine how exemplar choice affects the direction and cultural specificity of steered behavior.

**AI use** AI was used to debug the programming scripts, and Claude AI was used to produce plots and polish the final text of this manuscript. All ideas and errors are our own.

## B Related Work

This appendix provides an extended treatment of the related literature; the main text (Section 2) contains a condensed summary of the four gaps most directly addressed by STAR.

### B.1 Alignment and Social Behavior in LLMs

The alignment of large language models has evolved from instruction tuning (Ouyang et al., 2022) to reinforcement learning from human feedback (RLHF) (Li et al., 2023), with the goal of steering outputs toward helpful, harmless, and honest behavior. Instruction tuning fine-tunes a model on demonstration datasets to improve adherence to user intent, while RLHF introduces a reward model based on human preferences to guide optimization. These techniques have enabled even smaller aligned models to outperform larger unaligned counterparts in human evaluations.

More recent work extends alignment toward embedding broader societal values through curated datasets such as ETHICS (Hendrycks et al., 2023). However, aligning LLMs with high-level social traits, such as empathy and strategic politeness, remains challenging due to their context-sensitive and dynamic nature, which resists specification through static demonstrations or preference rankings. Despite progress, models frequently optimize superficial reward cues rather than internalizing intended values (Skalse et al., 2025), a failure mode characterized as the value-action gap (Shen et al., 2025). Such findings motivate interventionist methods that directly shape internal representations to achieve context-grounded, socially aligned behavior: critical in high-stakes settings such as coaching (Kosinski, 2024), negotiation (Street, 2024), and mental health support (Guo et al., 2024).

### B.2 Activation Steering and Representation Editing

A growing body of work investigates direct manipulation of internal representations via activation-based interventions. Causal tracing (Meng et al., 2023), activation patching, and contrastive activation addition (CAA) (Turner et al., 2024; Panickssery et al., 2024) locate and manipulate the internal circuits responsible for behavioral traits. Representation Engineering (RepE) (Zou et al., 2023) builds on these ideas with a top-down approach to shaping behaviors like helpfulness through population-level activation manipulation. These methods operate on the residual stream, which serves as a high-dimensional communication layer within transformers (Elhage et al., 2021), and offer a principled means of intervening on model behavior via steering vectors injected at specific locations.

While effective for coarse attributes such as sentiment or toxicity, most existing applications of activation steering address traits with relatively linear and context-independent structure. In contrast, socially grounded behaviors are highly dependent on interactional context and thus harder to isolate. Our work addresses this gap by integrating causal attribution patching (Kramár et al., 2024; Meng et al., 2023) with contrastive steering through behavioral exemplars to identify the specific layers and token positions where affective and relational traits emerge. This enables targeted and minimally disruptive interventions, expanding the scope of activation engineering to include complex social and emotional constructs, with implications for interpretable AI (Brundage et al., 2020; Bender et al., 2021).

### B.3 Linear Representations of Behavioral Concepts

The Linear Representation Hypothesis posits that many behavioral and semantic attributes, such as sentiment and toxicity, are encoded along linear directions in activation space (Park et al., 2024; Tigges et al., 2023; Panickssery et al., 2024; Nanda et al., 2023). Vector arithmetic over word embeddings and residual activations has been used to modulate these attributes, and such linear directions have been found to be consistent across model scales and families (Tigges et al., 2023; Nanda et al., 2023).

However, much of this work focuses on static or binary features. In contrast, behaviors like empathy and self-disclosure are dynamic, relational, and conditioned on conversational context, raising the question of whether such traits can also be captured through steerable linear representations. Our work extends this hypothesis by demonstrating that targeted behavioral examples can yield effective steering vectors for these complex behaviors. By applying directionally consistent interventions to contextually localized activations, we show that linearity can support not just semantic shifts, but also real-time modulation of high-level social strategies.

### B.4 Applications in Socially Grounded Dialogue

While recent studies confirm that behavioral traits such as sentiment or factuality can be encoded along linear directions in activation space (Turner et al., 2024; Zou et al., 2023; Park et al., 2024; Tigges et al., 2023; Panickssery et al., 2024; Meng et al., 2023), steering models toward richer, context-sensitive traits, such as empathy and negotiation, remains an open challenge. These traits are not easily reducible to single-word labels or sentiment scores; they require pragmatic reasoning, stylistic modulation, and context-dependent behavior.

This gap is particularly consequential in high-stakes applications. In mental health support, LLMs must establish emotional rapport and trust while minimizing harmful responses (Guo et al., 2024). In negotiation and coaching, success depends on interpreting emotional cues and adjusting strategies dynamically (Kosinski, 2024; Street, 2024). However, current models frequently fail to generate emotionally appropriate responses or sustain interpersonal trust, reflecting broader concerns about the opacity and unreliability of LLMs (Bender et al., 2021; Brundage et al., 2020). By enabling targeted interventions on internal activations, our method offers a more interpretable and controllable approach to social behavior modulation, advancing affectively reliable systems that bridge linguistic fluency and socially aligned interaction.

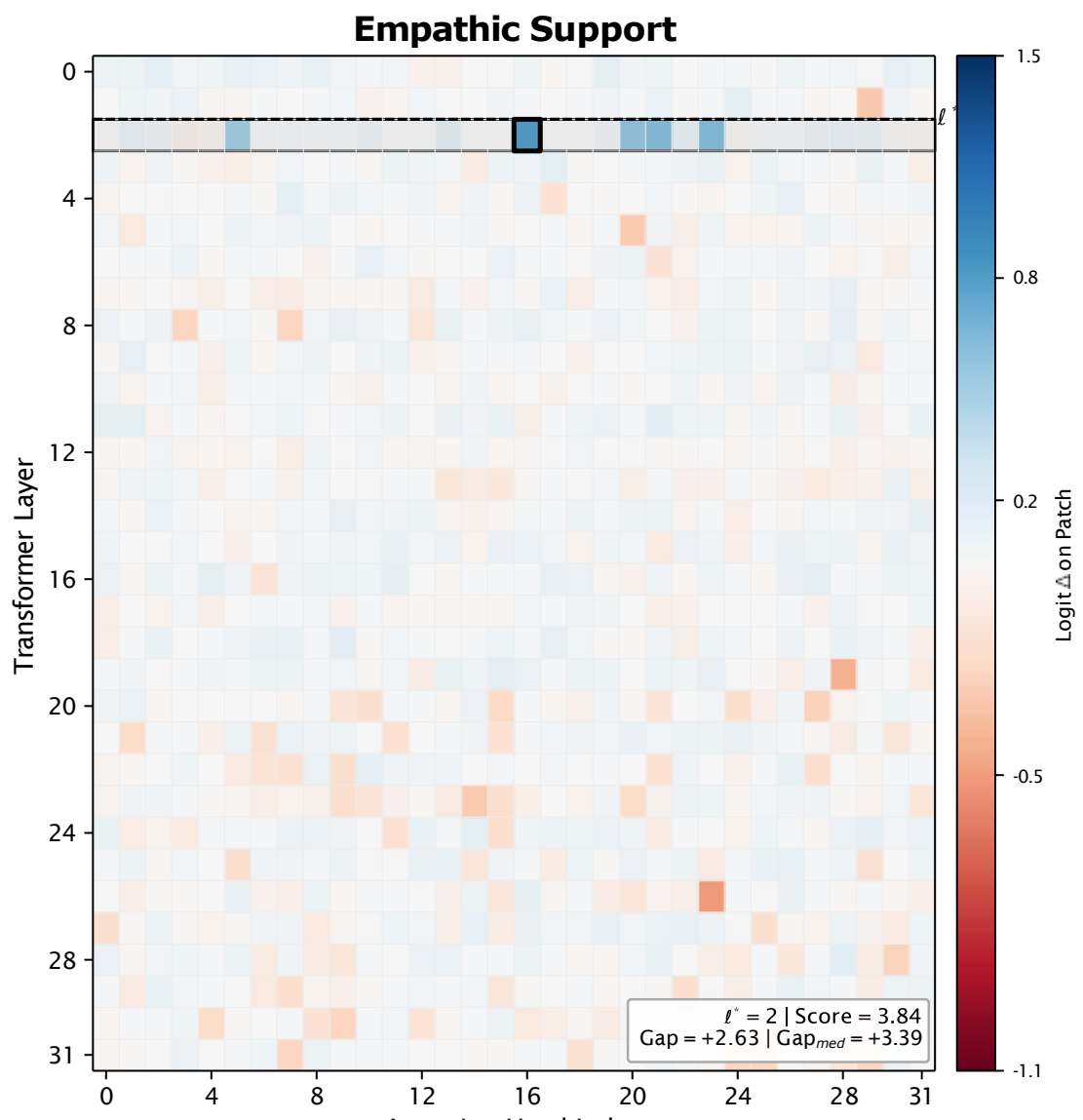

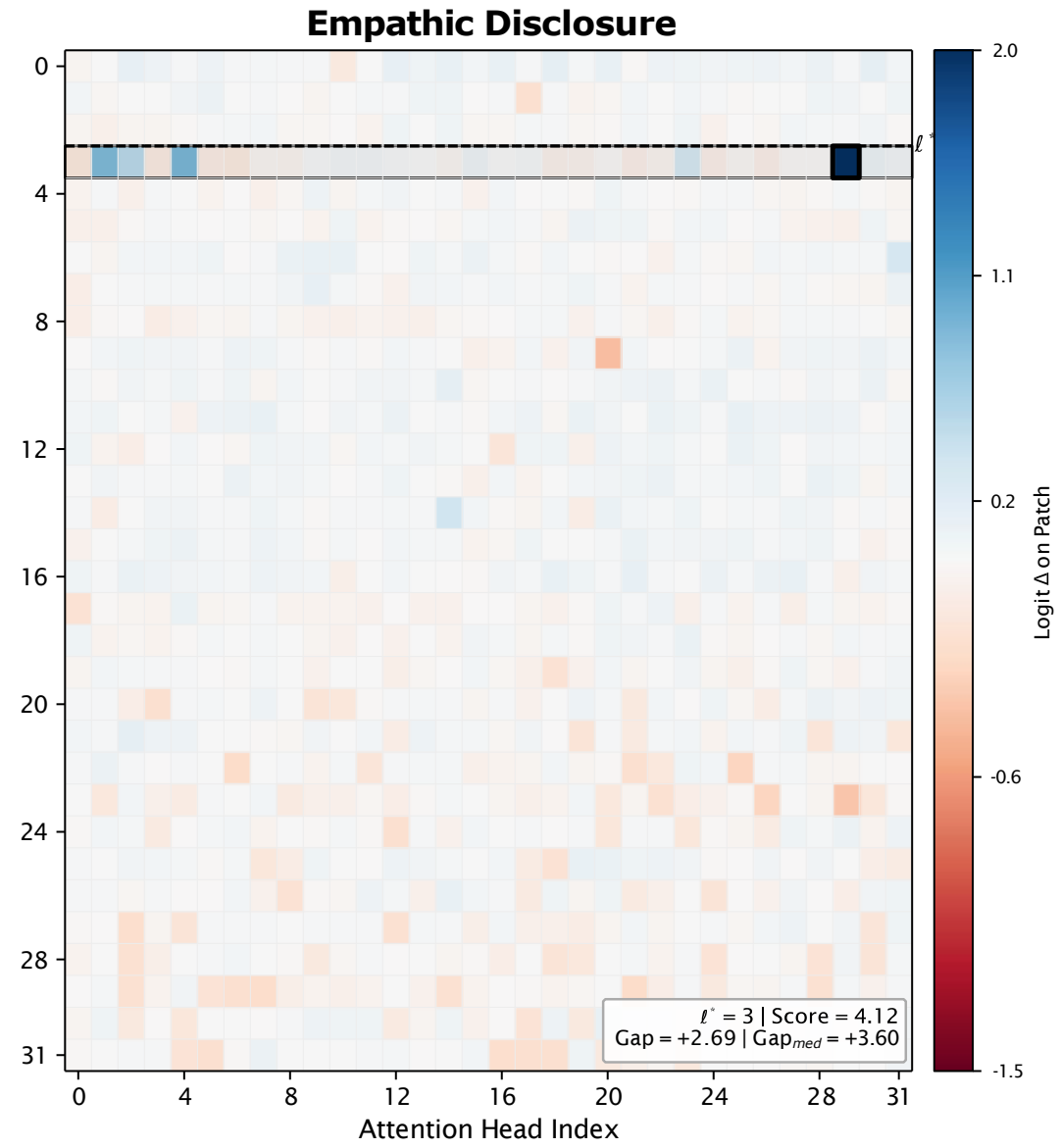

Figure 2: Attribution patching heatmaps for the two *emotional* diagnostic tasks. Each cell shows the causal attribution score (logit Δ on patch) for a single attention head (x-axis, 0-31) at a given transformer layer (y-axis, 0-31), measuring how much patching that head's output from a clean run into a misaligned run shifts the target logit toward the desired behavior. Blue indicates positive causal influence (patching improves alignment); red indicates opposing influence. The dashed horizontal line marks the identified intervention locus $\ell^*$; the black box highlights the peak-attribution head at that layer. Per-panel statistics (Score, Gap, $\text{Gap}_{\text{med}}$) are defined in Table 2. Early-layer heads ($\ell \leq 3$) carry concentrated positive influence, while mid-to-late layers exhibit diffuse opposing attribution, a pattern consistent across both tasks.

## C Attribution Patching: Methodology and Full Heatmaps

The central claim of STAR's first stage is that a single fixed injection point is insufficient for behavioral steering across task types: emotional and negotiation scenarios require intervention at qualitatively different layer depths and token positions. This section provides the mechanistic evidence for that claim via attribution patching across four diagnostic scenarios, **Empathic Support**, **Empathic Disclosure**, **Counter-Offer**, and **Strategic Politeness**, chosen to span qualitatively distinct communicative demands so that any consistent pattern cannot be attributed to a single surface-level cue.

**Logit Difference Metric.** The attribution score for a given activation is the change in a logit difference metric when that activation is replaced with its clean-run counterpart. The logit difference is defined as $\text{logit}(t^+) - \text{logit}(t^-)$, where $t^+$ is the target (desired) token and $t^-$ is a prominent undesired token, evaluated on the misaligned run after patching. In practice this is computed efficiently via a single backward pass on the misaligned prompt: gradients with respect to each activation are cached, and the attribution score is approximated as □ ((clean_act − misaligned_act) × grad_act). Positive scores indicate that the activation carries signal causally supportive of the desired behavior; negative scores indicate opposing influence.

**Procedure.** For each scenario we constructed paired clean/misaligned prompt completions (see Table 1 for exact prompts) and performed activation patching head-by-head across all 32 layers and 32 attention heads. We aggregate per-layer by summing head scores, and define the intervention locus $\ell^*$ as the layer with the highest aggregate positive attribution.

**Results.** Table 2 reports $\ell^*$, the cumulative attribution score at that layer (Score), its margin over the runner-up layer (Gap), and its margin over the median layer score ($\text{Gap}_{\text{med}}$). Figures 2 and 3 show the full 32×32 head-level heatmaps; Figures 4 and 5 show the corresponding layer-by-token-position heatmaps.

Two findings are consistent across all four tasks

**Negotiation — Attribution Patching: Layer-wise Head Contributions**

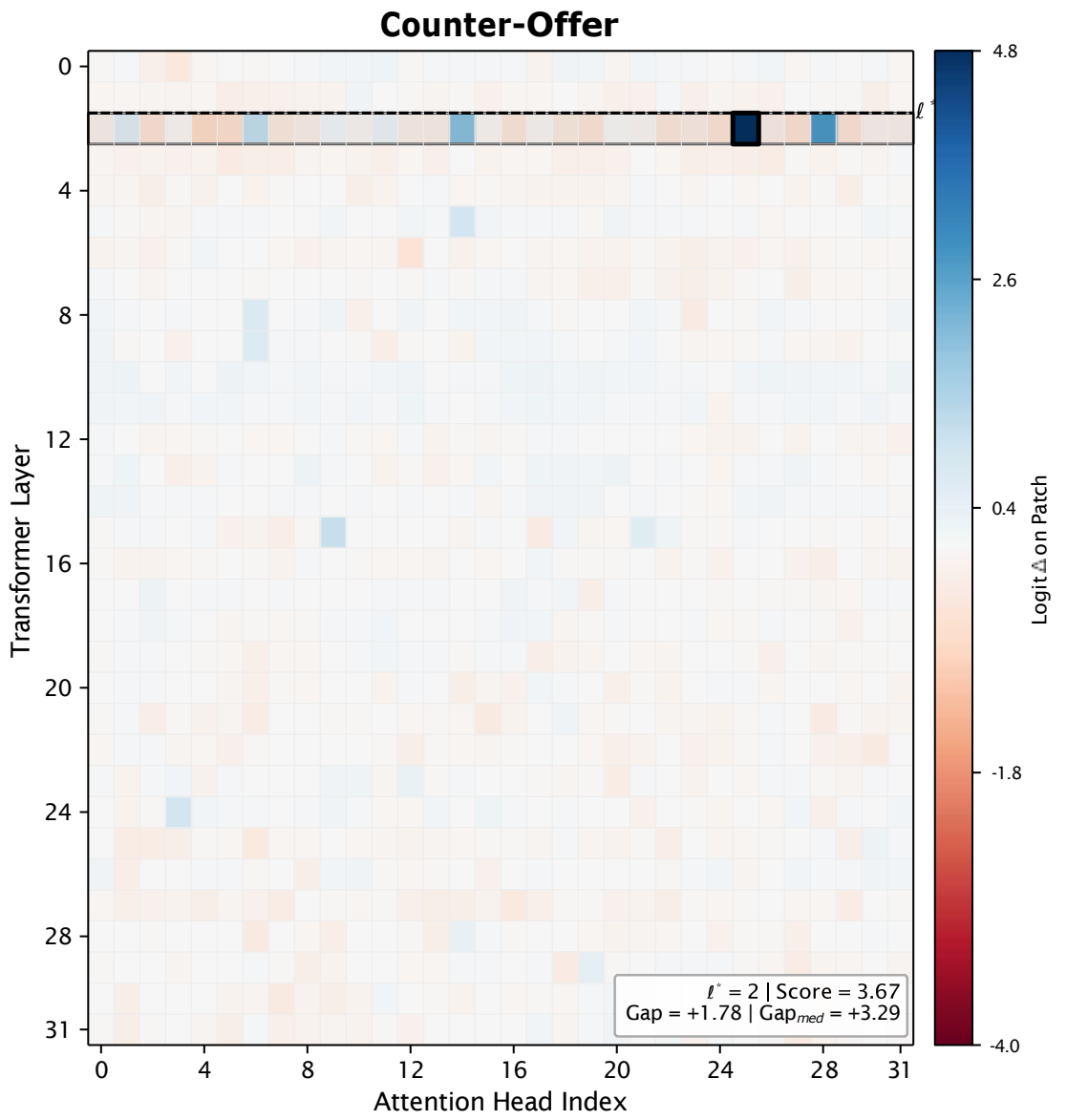

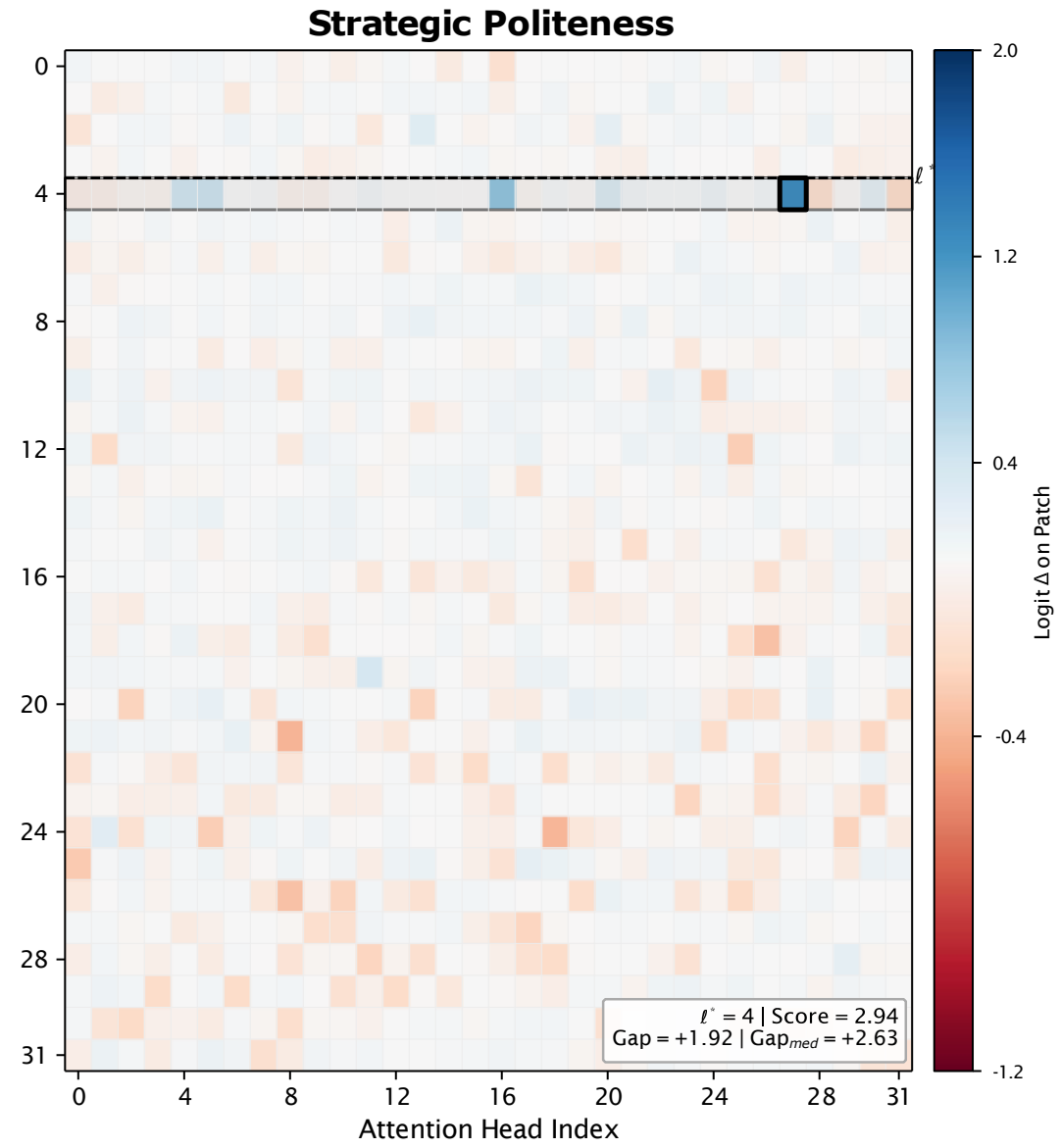

Figure 3: Attribution patching heatmaps for the two *negotiation* diagnostic tasks (same layout as Figure 2). Counter-Offer exhibits substantially larger absolute attribution scores (colorbar range $\approx \pm 4.8$) than Strategic Politeness ($\approx \pm 2.0$), suggesting that offer-acceptance behavior is more causally localised within early attention heads. Despite the difference in magnitude, the structural pattern holds: $\ell^* \in \{2, 4\}$ in both cases, with the peak head boxed at that layer. The contrast in scale across panels motivates per-task colorbars rather than a shared axis, which would wash out the Strategic Politeness signal.

and directly inform STAR's design. First, the identified locus is always an early layer ($\ell^* \in \{2, 3, 4\}$), preceding the mid-network layers typically associated with factual recall and reasoning. Second, early-layer heads exhibit concentrated positive attribution while late-layer MLP outputs show diffuse opposing attribution, a pattern characteristic of the model partially suppressing the target behavior in its default generation mode.

The token-position heatmaps reveal a structural dissociation between task types. Emotional tasks concentrate attribution in the final 15 prompt tokens and in shallow components (embed, layers 0-7), consistent with empathic generation being governed by the immediate conversational context. Negotiation tasks, by contrast, distribute attribution across multiple token clusters spanning the full prompt, with mid-layer MLPs dominant, consistent with bargaining behavior integrating information across conversational turns. This dissociation is the direct empirical motivation for running attribution patching per task rather than assuming a shared injection point.

## D Steering Vector Derivation and Application Details

This section details the pipeline through which STAR constructs and applies steering vectors, comprising three steps: behavioral exemplar construction, vector computation, and coefficient selection. Target layer identification (the prerequisite to all steps below) is derived via attribution patching and reported in Appendix C; we use those results directly here ($\ell^* = 2$ for Empathic Support, $\ell^* = 3$ for Empathic Disclosure, $\ell^* \in \{2, 4\}$ for negotiation tasks).

### D.1 Designing Steering Probes

Steering probes serve Stage 1 only: they are minimal, controlled dialogue contexts used to identify the layer-position pair ($\ell^*$, k) where each behavioral trait causally emerges via attribution patching.

Three structural constraints shaped the probe format:

- Completions must be *single tokens* to enable clean logit-difference computation via `to_single_token()`, and probes must be a total of 50 tokens long to allow full-context attribution patching across the prompt span.

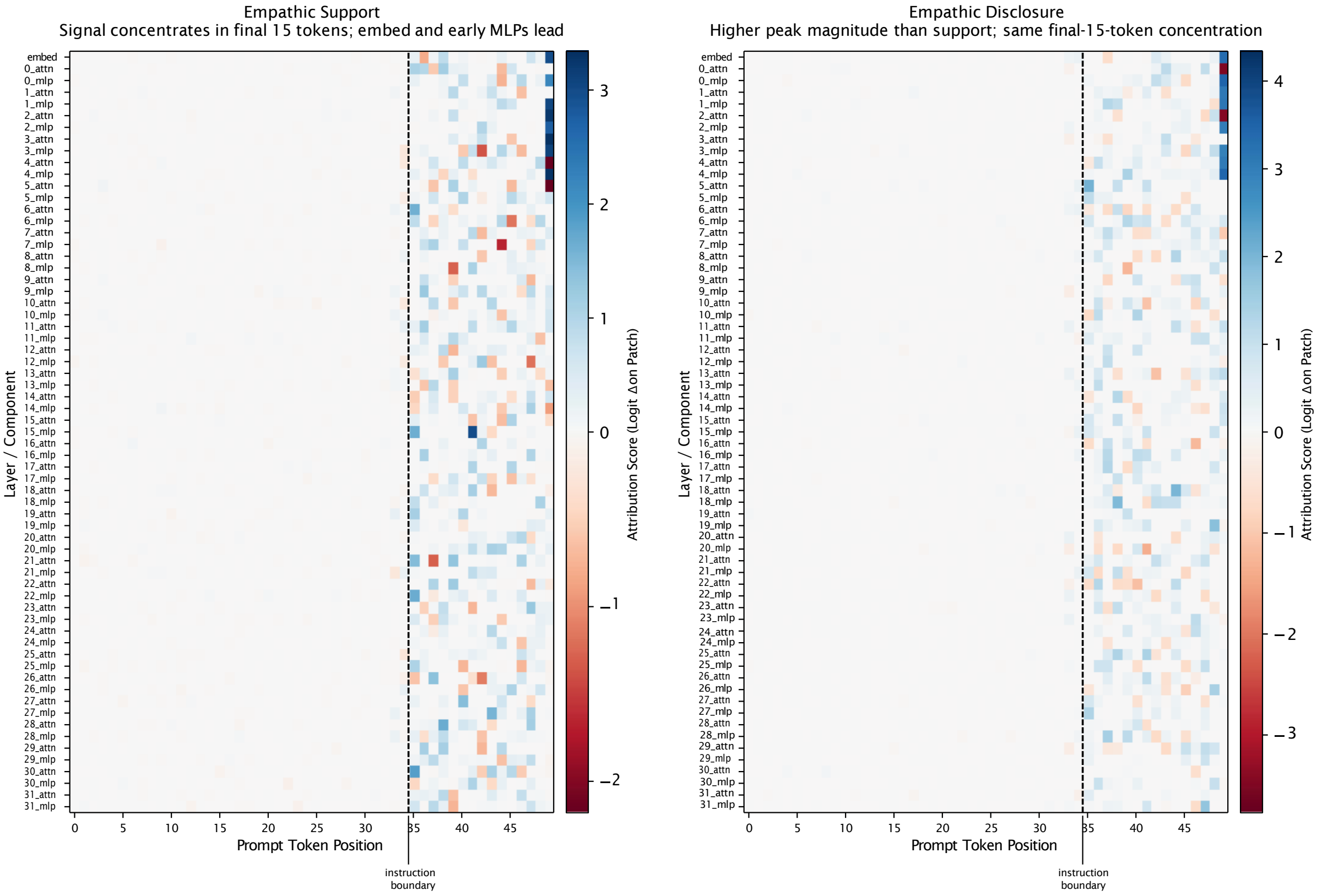

Figure 4: Attribution patching heatmaps for the two *emotional dialogue* diagnostic tasks (Empathic Support, left; Empathic Disclosure, right). Each cell reports the mean logit-Δ under activation patching for a given layer-component (rows: `embed`, then interleaved attention and MLP outputs for layers 0-31) and prompt token position (columns 0-49). The dashed vertical line marks the instruction boundary (token position 35), separating the system prompt from the dialogue turn; note that attributions concentrate almost entirely in the final 15 tokens, indicating that the model's empathic generation is governed by the immediate conversational context rather than earlier system-level tokens. Within that region, the embed layer and early MLP components (layers 0-7) carry the dominant positive signal. Empathic Disclosure (right) exhibits a higher peak magnitude (colorbar range ≈ ±4.3) than Empathic Support (≈ ±3.3), suggesting that self-disclosure behavior is more causally localised. Per-task colorbars are used throughout to avoid washing out the lower-magnitude Support signal under a shared axis.

Restricting answers to a single token isolates the model's preference between two candidate words without introducing multi-token decoding effects.

- Probe contexts must be generated synthetically but domain-grounded. This approach keeps probes short, controlled, and comparable across items. We instructed the model to produce short dialogue contexts reflecting the interactional style of two benchmark datasets: informal personal disclosures which mirror discussions found in communities such as *r/offmychest* and *r/CasualConversation*, and buyer-seller exchanges involving counter-offers, offer acceptances, and negotiations.
- Probes must follow a `Person A / Person B` framing to signal dyadic dialogue and anchor the completion in a conversational role and stabilizes the context across probes. This also distinguishes the task from open-ended or question–answer style completion.

Corrupted inputs are constructed by swapping each prompt with its paired neighbor, following standard attribution patching practice (Kramár et al., 2024). Generation was performed zero-shot using deterministic decoding (ChatGPT-5.3; temperature = 0, top_p = 1) with the following prompt:

> *Generate steering probes in the form of short dialogue contexts that elicit a single-token next-turn completion. Each probe should follow this format:*
>
> *Person A: [conversational turn]*
> *Person B:*

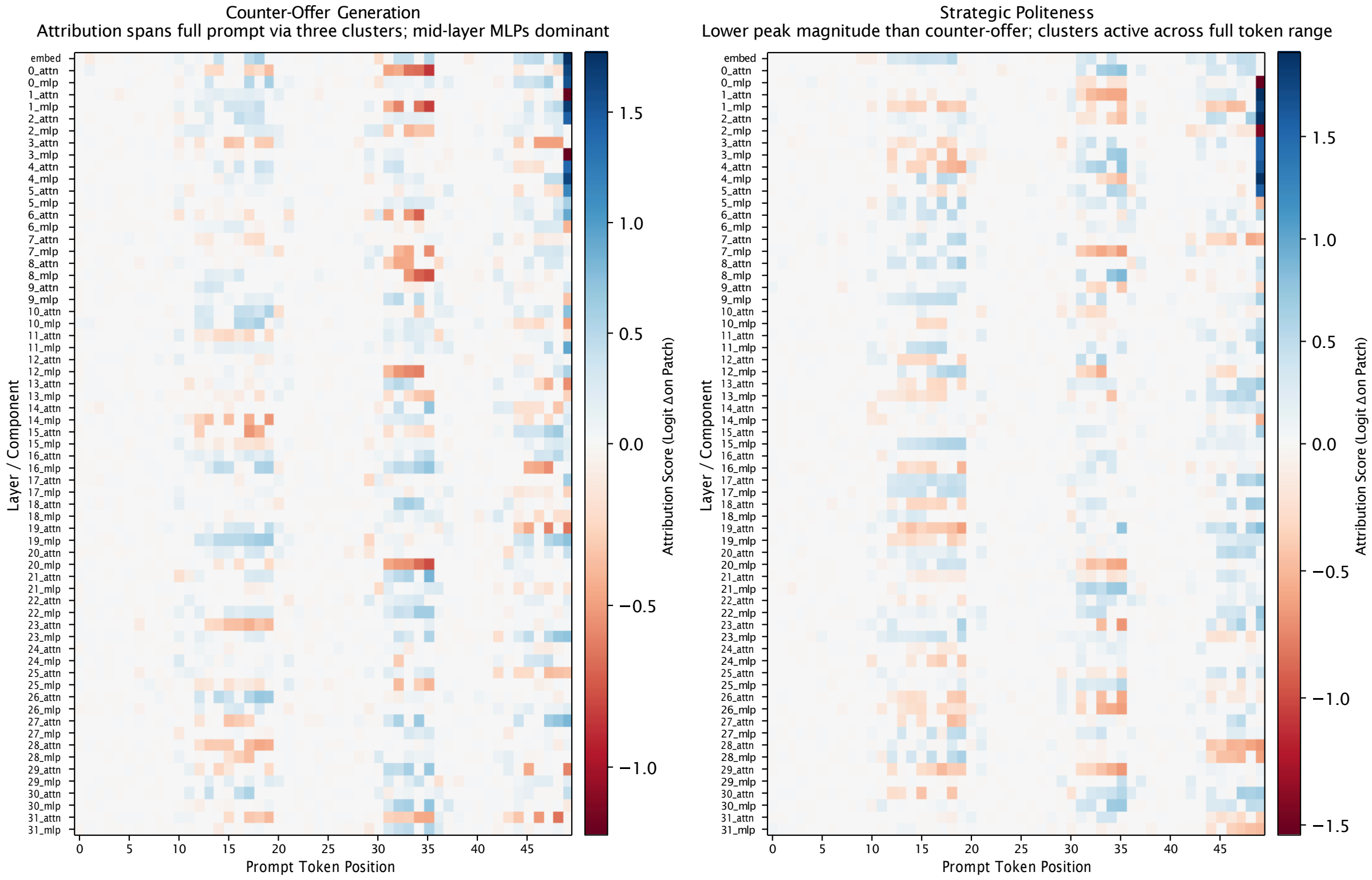

Figure 5: Attribution patching heatmaps for the two *negotiation* diagnostic tasks (Counter-Offer Generation, left; Strategic Politeness, right), using the same layout as Figure 2. Unlike the emotional tasks, attribution here is *not* boundary-gated: signal is distributed across the full prompt via three distinct token clusters (approximately positions 12-20, 31-36, and 44-50), consistent with the hypothesis that bargaining behavior integrates information from multiple conversational turns rather than a single decisive context window. Counter-Offer Generation exhibits substantially larger absolute attribution scores (colorbar range ≈ ±1.8) than Strategic Politeness (≈ ±1.9), with mid-layer MLP components carrying the dominant signal in both cases. The contrast between Figures 2 and 3 motivates a key claim of STAR: emotional and negotiation steering require injection at qualitatively different token positions and layer depths.

*Correct: [space-prefixed single word]*
*Incorrect: [space-prefixed single word]*

*Contexts: Generate 20 steering probes for each of the following domains. (1)* ***Conversational Dialogue****: Person A shares a personal disclosure or emotionally charged statement in casual conversation, similar to posts seen in r/OffMyChest or r/CasualConversation. Correct answers should express empathy (2)* ***Negotiation****: Person A makes a buyer offer or request in a casual marketplace setting. Correct answers should make a counter-offer.*

*Output as a Python dictionary with keys* `empathy` *and* `negotiation`*, each containing* `prompts` *and* `answers` *lists.*

*Rules: (1) Correct and incorrect answers must each be a single token. (2) Correct and incorrect answers must form clear semantic poles. (3) Incorrect answers must be unambiguously wrong in context. (4) Avoid incorrect answers that could plausibly fit the context. (5) All prompts must fit within 50 tokens.*

**From Probes to Behavioral Exemplars.** Stage 2 requires a separate input: a set of longer, naturalistic utterances capturing the behavioral direction to be injected. Unlike probes, these are not cloze completions but full conversational turns representative of the target trait ($D^+$) and its absence ($D^-$), together forming the *behavioral exemplar set* used to compute $V_{steer}$.

### D.2 Behavioral Exemplar Construction

For each behavioral dimension, behavioral exemplars were built as follows:

1. **Seed pair.** We began with a single manually

authored seed pair (one positive ($D^+$) and one negative ($D^-$) example) shown in Table 6. Positive examples reflect the target behavior (e.g., emotionally supportive, self-disclosing, or cooperative); negative examples reflect its absence (e.g., dismissive, impersonal, or uncooperative).

2. **GPT-4 augmentation.** GPT-4 was prompted to generate at least 10 additional examples per polarity class, conditioned on the seed pair and constrained to informal, personal conversational contexts: expressions of emotional distress, casual disclosures, and everyday interpersonal exchanges for the emotional tasks; buyer-seller marketplace exchanges for negotiation. Generated examples were manually reviewed to filter out borderline or ambiguous cases.
3. **Token normalization.** All texts were truncated or padded to a fixed token length before activation extraction. This ensures that mean activation differences between $D^+$ and $D^-$ reflect behavioral content rather than sequence-length confounds.

Final behavioral exemplars comprised at least 11 examples per polarity class per behavioral dimension (seed + ≥10 generated), for a total of ≥22 texts per steering vector. The seed pairs used to generate behavioral exemplars are reported in Table 6.

### D.3 Vector Computation

The model was run over each text in the behavioral pairs and residual-stream activations were extracted at the target layer $\ell^*$ across all token positions. Activations were averaged within each polarity class, yielding a mean positive tensor $\mu^+$ and a mean negative tensor $\mu^-$. The steering vector is:

$$\mathbf{v}_{\text{steer}} = \mu^+ - \mu^- \quad (5)$$

At inference, hidden activations at layer $\ell^*$ are modified by adding $\alpha \cdot \mathbf{v}_{\text{steer}}$ at the final 15 token positions, the region identified as carrying dominant causal attribution in Appendix C.

### D.4 Steering Coefficient Selection

**Emotional domain.** We swept $\alpha \in \{0.5, 1.0, 1.5, 2.0, 2.5, 3.0, 3.5, 4.0\}$ on a held-out validation set of N=100 BOLT SMS dialogues (sampled deterministically with seed 42, fixed prior to any sweep), measuring positive-sentiment probability (SST-2, ↑ better) as a proxy for emotional affect and completion-only perplexity (↓ better) as a proxy for fluency.

| Task | Polarity | Seed Text |
|---|---|---|
| Emotional Support | Positive (Supportive) | “That sounds really tough. I’m so sorry you’re dealing with this. I’m here to listen if you want to talk more.” |
| | Negative (Dismissive) | “Okay, that event occurred. Let us look at it rationally. What is the logical next action you should consider taking now?” |
| Emotional Disclosure | Positive (Disclosing) | “To be honest, I’ve been feeling quite stressed and uncertain about things lately. It’s been weighing on me.” |
| | Negative (Impersonal) | “I generally prefer to keep my personal feelings to myself. As for work, everything is proceeding according to plan.” |
| Counter-Offer | Positive (Cooperative) | “I understand where you’re coming from with that price. Would you be open to meeting somewhere around $70?” |
| | Negative (Uncooperative) | “That price doesn’t work for me at all. My offer is $60 and I’m not going any higher.” |
| Strategic Politeness | Positive (Validating) | “I understand why you priced it that way based on the condition. That makes sense from your perspective.” |
| | Negative (Dismissive) | “I don’t really care about that justification. The price is still way too high.” |

Table 6: The pairs used in Stage 2 to construct behavioral exemplar vectors $V_{\text{steer}}$ for each behavioral target. Positive and negative examples define $D^+$ and $D^-$ respectively.

Affect increases monotonically from 0.65 at α=0.5 to a peak of 0.96 at α=2.5, while perplexity rises from 20 to 90 over the same range and continues to 450 at α=4.0. A marginal efficiency analysis (normalised Δaffect/ΔPPL between consecutive steps) reveals a sharp cost-benefit collapse at α=2.0 → 2.5: affect improves by only 0.04 (+4.3%) while perplexity increases by 35 points (+64%). Beyond α=2.5, affect declines while perplexity continues to rise. We therefore selected α=2.0 as the last Pareto-efficient operating point, where affect reaches 0.920 ± 0.007 at a perplexity of 55, a 1.85× improvement over the unsteered baseline (affect = 0.500, PPL = 18) with moderate fluency cost. This selection was pre-registered before test-set evaluation; see Figure 6.

**Negotiation domain.** A parallel sweep was conducted on N=100 held-out Craigslist Bargains dialogues, fixed prior to any sweep. Agreement rate increases from 0.41 at α=0.5 to a peak of 0.78 at α=2.5, while repetition score degrades sharply beyond that threshold. The same marginal efficiency criterion identifies α=2.5 as the Pareto-efficient operating point, with semantic coherence remaining stable up to that value before degrading. All negotiation experiments use α=2.5.

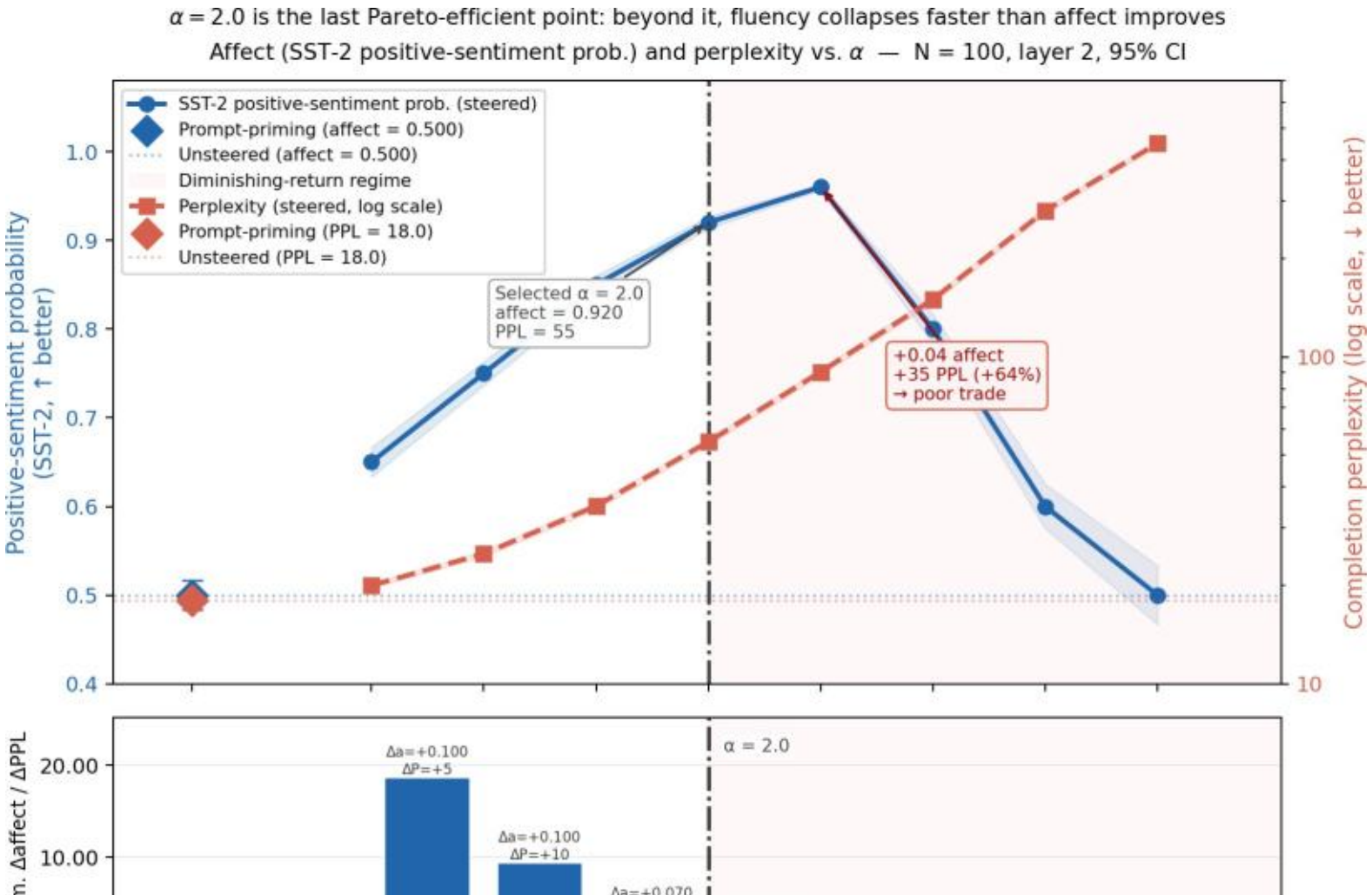

Figure 6: **Steering coefficient selection for the emotional domain.** *Top:* SST-2 positive-sentiment probability (blue, left axis) and completion perplexity (red, right axis, log scale) across $\alpha \in \{0.5, \ldots, 4.0\}$, evaluated on N=100 held-out BOLT SMS dialogues at layer $\ell^*=2$. Shaded bands denote 95% CI. The prompt-priming baseline (diamond markers) and unsteered condition (dotted lines) both sit at affect = 0.500, PPL = 18. *Bottom:* Marginal efficiency (Δaffect/ΔPPL, normalised) between consecutive α steps. α=2.0 fixed before any test-set evaluation.

## E Experimental Design and Evaluation

The design choices documented here directly support STAR's two central claims: that localized activation steering produces behaviorally coherent outputs across single-turn and extended multi-turn contexts, and that attribution-guided injection outperforms both instruction-level and unlocalized steering baselines. Domain-specific choices are documented separately below; all shared implementation details follow Section 3.6.

### E.1 Emotional Support and Disclosure

**Single-Turn Design.** For each dialogue in the BOLT SMS set, the conversational history up to Person B's response was used as the prompt. Two responses were generated per prompt: one unsteered (baseline) and one steered (with activation intervention at the final 15 token positions of the target layer). Evaluation focused on linguistic and emotional features of the single generated utterance; metrics are described in Appendix F.1.

**Multi-Turn Design.** The multi-turn experiment tested whether steered emotional behavior persisted, decayed, or could be reversed over extended interaction. Due to computational cost, this was conducted on the final 10% of eligible dialogues (1,102 examples). Each dialogue followed a three-exchange loop:

1. Person A (Mistral) generates a response to the initial human context.
2. Person B (Llama) generates a response (first steering point).
3. Person A (Mistral) generates a response.
4. Person B (Llama) generates a response (second steering point).
5. Person A (Mistral) provides a closing response.

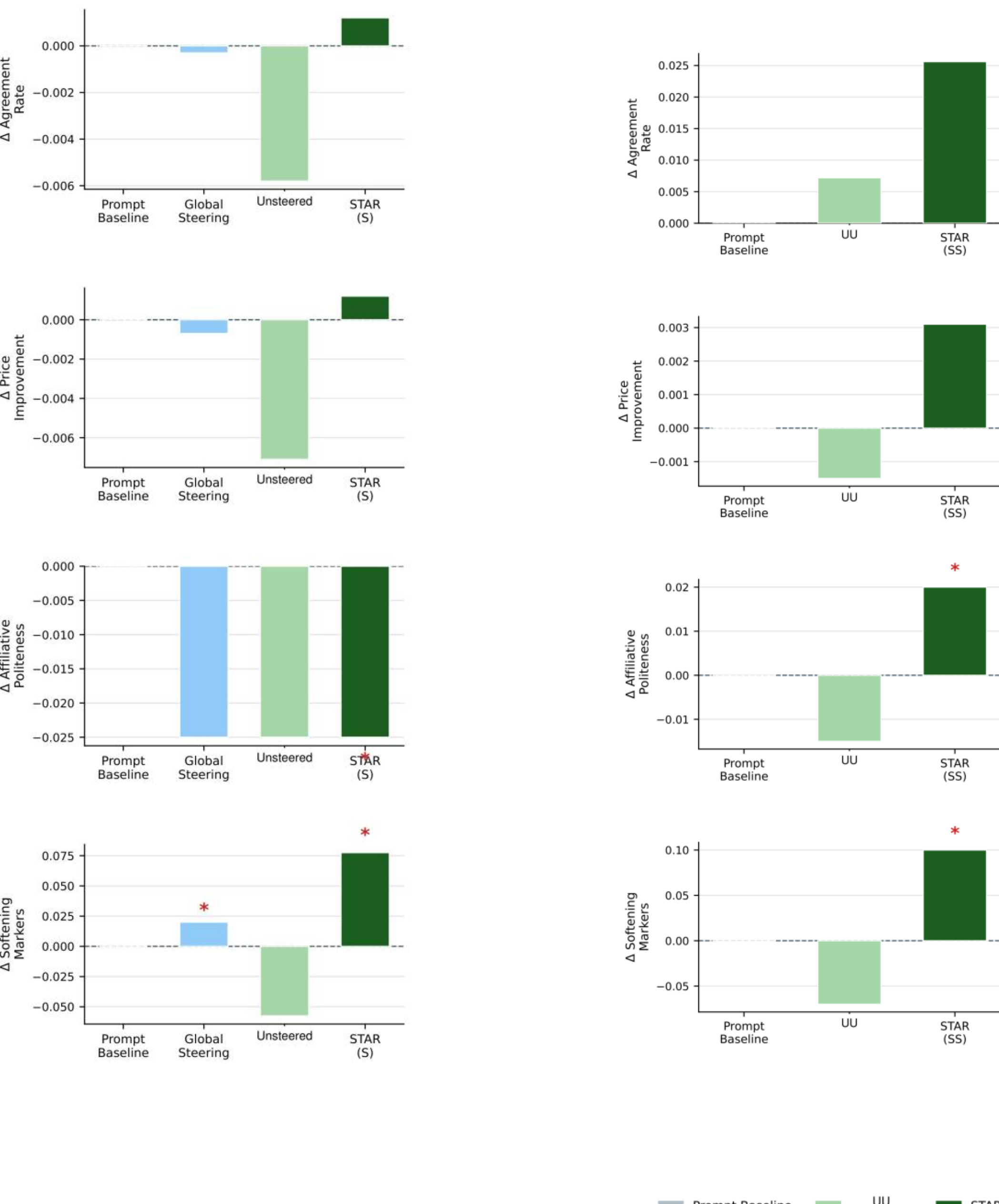

Figure 7: Metric shift from Prompt Baseline (Δ=0, dashed) for the single-turn Craigslist Bargain setting (n=5,155). Each bar shows Δ relative to the Prompt Baseline for four outcome categories: Agreement Rate (binary negotiation success), Price Improvement (% change from dataset price to simulated price), Affiliative Politeness (Gratitude proportion), and Softening Markers (mean of Hedges and Apologizing proportions; ConvoKit). Conditions: Prompt Baseline (grey), Global Activation Steering (light blue), Unsteered (light green), STAR (S) (dark green). * $p_{adj} < 0.05$ vs. Prompt Baseline (Benjamini-Hochberg correction).

Figure 8: Metric shift from Prompt Baseline (Δ=0, dashed) for the multi-turn Craigslist Bargain setting (n=515). Each bar shows Δ relative to the Prompt Baseline for the same four outcome categories as Figure 7: Agreement Rate, Price Improvement, Affiliative Politeness (Gratitude proportion), and Softening Markers (mean of Hedges and Apologizing proportions). Conditions: Prompt Baseline (grey), UU (light green), STAR (SS) (dark green). * $p_{adj} < 0.05$ vs. Prompt Baseline (Benjamini-Hochberg correction).

Mistral was conditioned via the following fixed system prompt throughout:

> *"You are Person A, a friend having a conversation with Person B. You are coming to them with a genuine problem or feeling of uncertainty. Your goal is to express your feelings honestly and see how they respond. You are not looking for simple advice or a quick fix, but rather for a sense of connection and understanding. Your responses should be natural and reflect your emotional state. React dynamically to Person B's tone: if they are supportive and empathetic, you can share more details about your situation; if they are dismissive, cold, or overly logical, you might become more reserved or express confusion."*

Four steering configurations were evaluated across this loop (identical to the negotiation multi-turn design below):

- **UU**: No steering at either turn.
- **US**: Steering introduced at the second turn only.
- **SU**: Steering applied at the first turn, removed at the second.
- **SS**: Steering maintained across both turns.

Evaluation was conducted at both the turn level and across aggregated Person B responses to assess consistency, escalation, and cumulative behavioral change.

### E.2 Negotiation Experiments

**Dataset and Preprocessing.** Experiments used the CraigslistBargain dataset (He et al., 2018), filtering for dialogues with at least 5 turns and at least one seller concession: a turn in which the seller lowers their previously stated price. Simulations were initialized from the dialogue context up to and including the buyer's response following this concession, ensuring that steering effects were measured in contexts where bargaining dynamics were already active.

**Single-Turn Design.** Conducted on all 5,155 valid dialogues. Mistral-7B-Instruct served as the Seller; steered Llama-3.1-8B served as the Buyer. The steering vector corresponded to the counter-offer diagnostic task from attribution patching (Appendix C), targeting offer-generation behavior at $\ell^* \in \{2, 4\}$. The goal was to measure the immediate effect of steering on the buyer's first counter-offer, evaluated on the following outcome metrics:

- **Agreement Rate.** Binary negotiation success, identified via keyword matching in the seller's final turn. This captures whether steering produced offers the seller was willing to accept, rather than merely shifting surface-level language.
- **Price Improvement.** Percentage change between the dataset's recorded final price and the final negotiated price in the simulation. This captures the economic direction of steering, distinguishing vectors that push toward better buyer outcomes from those that merely increase verbosity.

The seller's response to the steered or unsteered buyer turn was then used to assess acceptance or rejection.

**Multi-Turn Design.** Conducted on the final 10% of negotiation dialogues (515 examples). The loop structure was identical to the emotional multi-turn design above. Mistral was conditioned via the following fixed system prompt:

> *"You are the Seller in a negotiation on a marketplace. Your goal is to sell your item for the best possible price, while still being a reasonable and fair negotiating partner. You have a starting price in mind but are open to some negotiation; do not accept extremely low offers immediately. Your behavior should be principled but pragmatic. Respond directly to the Buyer's tone and strategy: if the Buyer is polite and provides good reasons for a lower price, you can make a sensible counter-offer. If the Buyer is rude, dismissive, or makes unreasonable demands, you should remain firm on your price or state that a deal may not be possible. Your aim is to reach a mutually agreeable deal, but not at a price that feels unfair to you."*

The same four UU/US/SU/SS configurations were evaluated, with the same turn-level and dialogue-level aggregation as the emotional setting.

**On the modest STAR vs. global steering distinction in negotiation.** The GS vs. LS comparison (Table 12, Comparison 3) shows trivial effect sizes on four of six categorical features, in contrast to the clearer advantage of localized over global injection in the emotional domain. A complementary finding in Table 13 (Single-Turn Comparison 2) is that global steering itself converges with the prompt baseline on most negotiation features, suggesting that unlocalized injection replicates prompt-level distributional effects rather than engaging internal representations in a meaningfully distinct way. We interpret both patterns as a consequence of the distributed attribution structure identified in Figure 5: unlike emotional tasks, where causal signal concentrates in the final 15 prompt tokens, negotiation attribution spans the full prompt via three token clusters. Global injection inadvertently covers the causally relevant positions, partially replicating localized injection; the residual advantage of STAR on Hedges and Apologizing ($V = 0.11$-$0.12$) reflects the precision benefit at the specific layers attribution patching identifies. This constitutes a meaningful boundary condition: localization provides the greatest advantage when causal attribution is sparse and positionally concentrated.

### E.3 Baseline Methodology

Three baseline conditions were implemented, as described in Section 3.6. We document the full prompt templates here for reproducibility.

**Unsteered Generation.** Llama-3.1-8B was run with identical decoding parameters (greedy decoding with repetition penalty) and no intervention of any kind, providing the absolute behavioral floor.

**Zero-Shot Instruction Priming.** A task-specific system message was prepended to the input, shaping behavior through language-level control alone without access to internal representations.

For negotiation:

> *"You are a strategic and skilled negotiator acting as the buyer in a marketplace transaction. Your primary objective is to secure the item at the lowest possible price, demonstrating savvy and effective bargaining tactics. You must maintain a consistently polite and professional tone throughout the interaction, even when being firm. Your strategy is to be proactive. Justify your offers with clear reasons, such as the item's perceived condition, market comparisons, or logistical benefits you offer (like quick pickup). When the seller makes a counter-offer, acknowledge their position but always steer the conversation back towards a better deal for you. Use persuasive language, ask clarifying questions to gather information, and be prepared to walk away if the terms are not favorable. Your success is measured not just by the final price, but by the intelligence and civility of your negotiation strategy."*

For emotional support and disclosure:

> *"You are an emotionally intelligent and supportive conversational partner. Your primary function is to provide comfort, validation, and a safe space for the other person to express themselves. You must maintain a natural and empathetic conversation flow by asking thoughtful, open-ended questions, actively listening to their concerns, and responding in a way that shows you understand and care. Avoid giving generic advice or making abrupt topic changes; instead, focus on being present and supportive to encourage a connected and genuine emotional dialogue."*

**Global Activation Steering.** As defined in Equation 4, the same $\mathbf{v}_{\text{steer}}$ was applied across all layers and token positions without attribution-guided localization. This baseline held the vector constant while removing the localization that constitutes STAR's core contribution.

## F Evaluation Metrics

Automatic and human evaluation metrics jointly cover what neither modality can establish alone: automatic metrics provide scale and statistical power across thousands of examples, while human evaluation verifies that the distributional shifts captured automatically correspond to genuinely improved communicative quality as perceived by readers.

### F.1 Automatic Evaluation Metrics

Steered and unsteered outputs were evaluated at the utterance level using categorical and continuous linguistic features.

IAA STABILITY · ROLLING KAPPA ACROSS 200-ITEM ANNOTATION SESSION (50-ITEM WINDOWS)

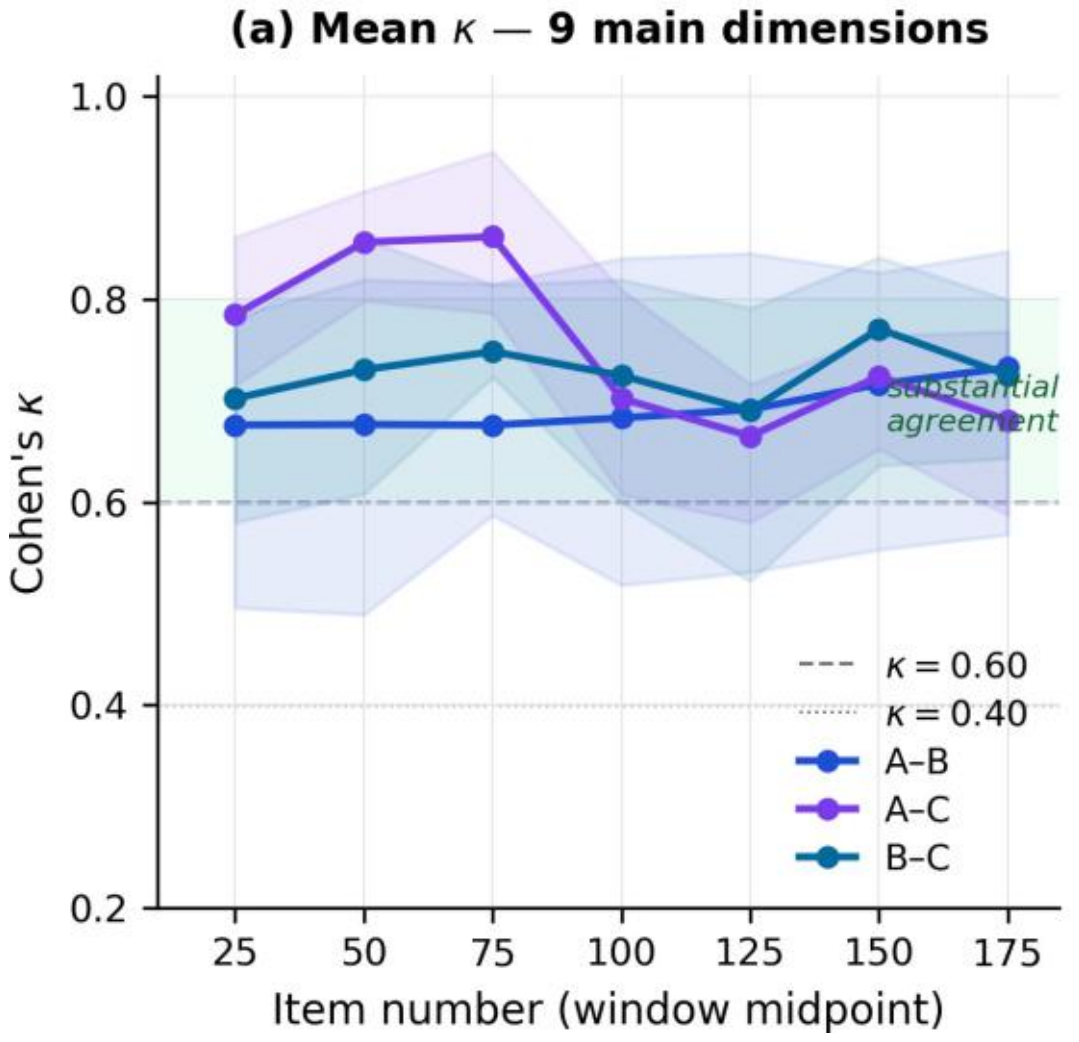

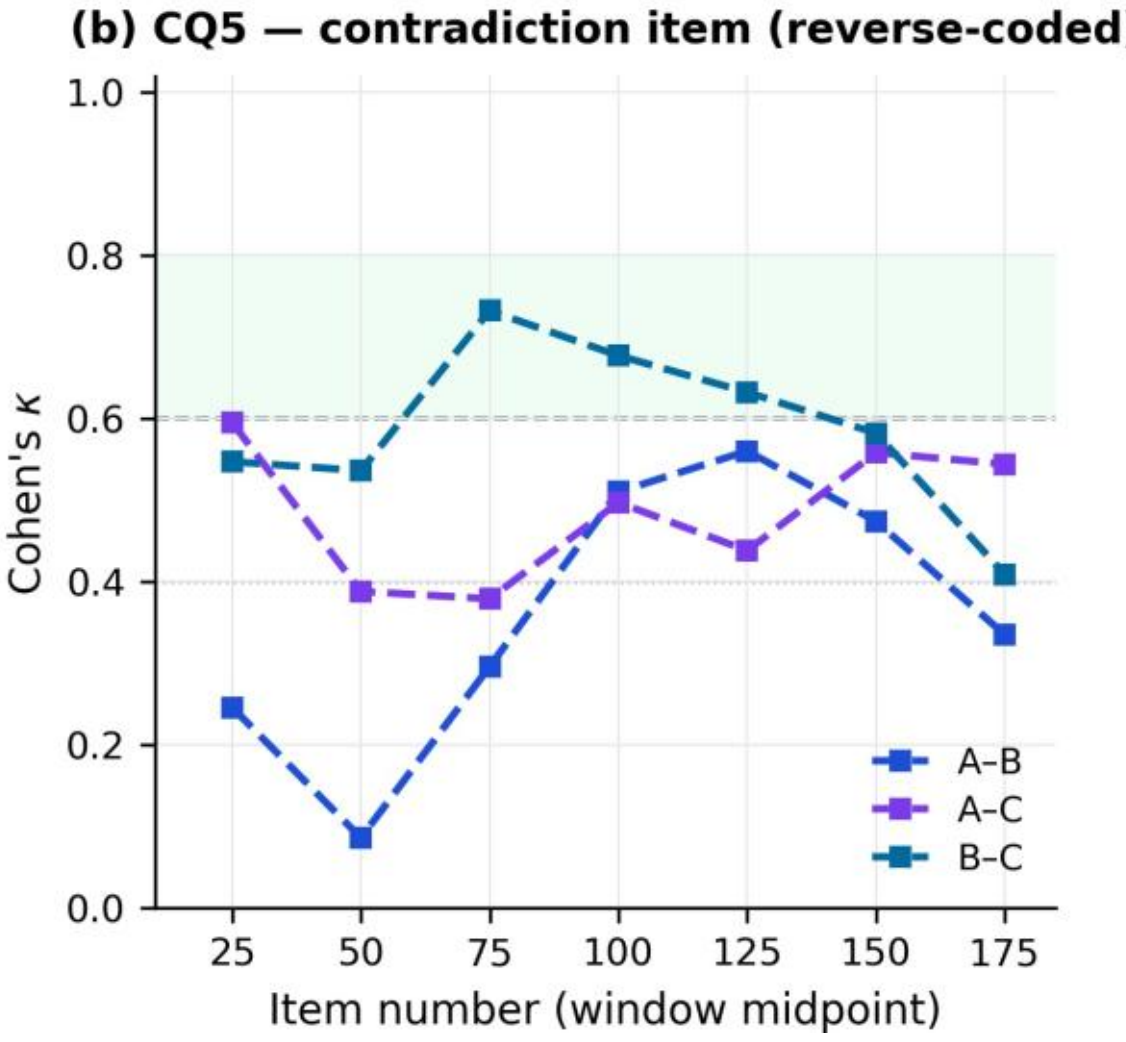

Figure 9: Rolling Cohen's κ across the 200-item empathy annotation session, computed in overlapping 50-item windows (x-axis = window midpoint), shown separately for each annotator pair (A–B, A–C, B–C). **(a)** Mean κ across the nine main dimensions (excluding CQ5); shaded bands indicate the per-window range across dimensions. Agreement remains stable throughout the session for all three pairs, with no evidence of fatigue-driven degradation, and all pairs remain above the substantial-agreement threshold (κ = 0.60, dashed line) at every window. **(b)** CQ5 (*Contains internal contradictions*, reverse-coded) shown in isolation. Unlike the nine main dimensions, CQ5 exhibits high volatility and drops below the moderate-agreement threshold (κ = 0.40, dotted line) at several windows, particularly for the A–B pair, confirming that CQ5 is an inherently harder item and motivating the interpretive caution applied to the Content Quality composite.

**Affective Outcomes.**

- **Sentiment Polarity.** Overall sentiment (positive vs. negative) was assessed using two complementary approaches. In the multi-turn setting, responses were classified using a DistilBERT model fine-tuned on SST-2 (Sanh et al., 2019), yielding a binary positive/negative label per utterance. In the single-turn setting, sentiment was operationalized as the net positive score: the proportion of positive NRC Emotion Lexicon tokens minus the proportion of negative tokens (Fast et al., 2017).

- **Politeness Strategies.** Categorical features denoting the presence of politeness markers were extracted using ConvoKit's `PolitenessStrategies` module (Chang et al., 2020). For the emotional domain these included: Gratitude, Apologizing, Hedges, and overall positive/negative tone markers. For the negotiation domain these additionally included: directness, indirect requests, and dismissiveness.

- **Empath Lexical Categories.** Continuous measures of empathy-relevant language were derived from the Empath lexicon (Fast et al., 2017), covering seven categories: *help*, *communication*, *speaking*, *listen*, *strength*, *healing*, and *nervousness*. Per-utterance scores were averaged across categories to yield a single empathy index.

**Negotiation Outcomes.**

- **Agreement Rate.** Binary negotiation success, identified via keyword matching in the seller's final turn. Captures whether steering produced offers the seller was willing to accept, rather than merely shifting surface-level language.

- **Price Improvement.** Percentage change between the dataset's recorded final price and the final negotiated price in the simulation. Captures the economic direction of steering, distinguishing vectors that push toward better buyer outcomes from those that merely increase verbosity.

Differences in categorical distributions between conditions were assessed using $\chi^2$ tests of independence. Continuous metrics were compared using Welch's t-tests. To control for Type I error inflation across multiple feature tests, p-values were adjusted using the Benjamini-Hochberg FDR procedure; adjusted p-values below 0.05 were considered statistically significant.

| Dim. | Item | % Agr. | $\kappa$ A–B | $\kappa$ A–C | $\kappa$ B–C | ICC2k |
|---|---|---|---|---|---|---|
| *Content Quality* | | | | | | |
| | CQ3 | 71.5 | 0.788 | 0.766 | 0.773 | **0.925** |
| | CQ5† | 67.5 | 0.365 | 0.513 | 0.583 | 0.735 |
| | CQ6 | 68.5 | 0.735 | **0.824** | 0.760 | **0.926** |
| *Contextual Fit* | | | | | | |
| | CF1 | 71.0 | 0.777 | 0.784 | 0.758 | 0.921 |
| | CF2 | 57.5 | 0.614 | 0.677 | 0.649 | 0.867 |
| *Emotional Quality* | | | | | | |
| | EQ1 | 63.0 | 0.739 | 0.759 | 0.744 | 0.924 |
| | EQ2 | 60.0 | 0.727 | 0.752 | 0.738 | 0.922 |
| | EQ3 | 58.0 | 0.661 | 0.708 | 0.674 | 0.888 |
| *Perceived Naturalness* | | | | | | |
| | PN3 | 65.0 | 0.575 | 0.718 | 0.635 | 0.842 |
| | PN4 | 70.5 | 0.664 | 0.757 | 0.735 | 0.886 |
| *Cronbach's $\alpha$* | A: 0.934 | B: 0.969 | C: 0.944 | | | |

Table 7: Inter-annotator agreement for the empathy human evaluation (N = 200 items, three annotators: A, B, C). **% Agr.** = three-way percentage agreement. $\kappa$ = Cohen's kappa for each annotator pair. **ICC2k** = average-measures ICC2 (k = 3); values ≥ 0.75 indicate good reliability. **Bold** marks the highest value in each column. † CQ5 (*Contains internal contradictions*, reverse-coded) was the weakest dimension across all metrics; its lower ICC2k (0.735) reflects inherent ambiguity in detecting logical contradiction in short conversational turns. Cronbach's α computed across all ten items per annotator confirms strong internal consistency.

| Dimension | Training<br>*N = 75* | Empathy<br>*N = 200* | Negotiation<br>*N = 200* |
|---|---|---|---|
| CQ3 | 0.841 | **0.925** | **0.925** |
| CQ5† | 0.633 | 0.735 | **0.923** |
| CQ6 | 0.743 | 0.926 | **0.927** |
| CF1 | **0.848** | 0.921 | **0.926** |
| CF2 | 0.810 | 0.867 | **0.911** |
| EQ1 | 0.821 | **0.924** | 0.899 |
| EQ2 | 0.746 | **0.922** | 0.889 |
| PN3 | 0.746 | 0.842 | **0.880** |
| PN4 | 0.780 | 0.886 | **0.901** |
| *Mean* | 0.774 | 0.883 | **0.911** |

Table 8: ICC2k (average-measures) across the three annotation phases. **Training** (N = 75) was a calibration round conducted before the main empathy annotation. **Empathy** (N = 200) and **Negotiation** (N = 200) are the two final annotation tasks. **Bold** marks the highest value per row. Mean ICC2k rose from 0.774 in training to 0.883 after calibration, confirming that annotator agreement improved measurably across the annotation process. † CQ5 remains the weakest dimension throughout all three phases.

### F.2 Human Evaluation Protocol

Human evaluation used a structured annotation framework synthesized from prior work on dialogue quality assessment (See et al., 2019; Rashkin et al., 2019; Pennycook and Rand, 2019), organized into four dimensions. Each dimension targets a distinct failure mode that automatic metrics cannot reliably detect: distributional sentiment shifts do not guarantee grammatical fluency, cross-turn coherence, proportionate emotional tone, or the absence of tell-tale automated phrasing. The empathy and negotiation tasks were annotated in separate exercises with independent training phases; IAA results are reported for each separately below.

**Sampling.** We sampled 200 items per domain for annotation, stratified across steering conditions to ensure representation of each experimental configuration. Items were drawn from the full evaluation sets (empathy: n=11,002 single-turn dialogues; negotiation: n=5,155 single-turn dialogues) using a fixed random seed. Per-condition sample sizes ranged from N=10 to N=46 for both the empathy and negotiation tasks, reflecting the unequal distribution of steering conditions across the sampled pool.

**Annotators and training.** Three annotators participated: A and B served as primary raters, and C served as the reference rater and adjudicator throughout. Prior to the main empathy annotation, all three completed a 75-item calibration round drawn from a held-out subset of the negotiation task. A and B rated independently, then reviewed disagreements against C's ratings to establish shared anchoring on the scale endpoints before proceeding to the main task. Agreement improved measurably across phases: mean ICC2k rose from 0.77 at calibration to 0.88 on the empathy task and 0.91 on the negotiation task (Table 8), confirming that the training protocol achieved its intended effect. Annotators were blind to condition labels throughout.

**Adjudication.** Final scores were determined by majority vote across all three annotators (A, B, and C). For each item and dimension, the score held by at least two of the three annotators was taken as the final value. On items where all three annotators assigned different scores— rendering a strict majority impossible—the final score was set to the median of the three ratings, rounded to the nearest integer.

| Dimension | Single-turn ($N$ = 88) | Multi-turn ($N$ = 112) |
|---|---|---|
| *Content Quality* | | |
| CQ3 | **0.895** | 0.699 |
| CQ5† | 0.395 | 0.342 |
| CQ6 | **0.798** | 0.688 |
| *Contextual Fit* | | |
| CF1 | **0.834** | 0.729 |
| CF2 | **0.696** | 0.547 |
| *Emotional Quality* | | |
| EQ1 | **0.813** | 0.680 |
| EQ2 | **0.796** | 0.671 |
| EQ3 | **0.706** | 0.627 |
| *Perceived Naturalness* | | |
| PN3 | **0.639** | 0.517 |
| PN4 | **0.700** | 0.636 |
| *Mean (excl. CQ5)* | **0.764** | 0.644 |

Table 9: Cohen's κ (annotator pair A–B) by turn type for the empathy annotation. Single-turn items achieve substantially higher agreement across every dimension (mean κ = 0.764 vs. 0.644, excluding CQ5), consistent with single-turn responses providing an unambiguous reference context for each judgement. Lower multi-turn agreement reflects interpretive variation about which turn a judgement should anchor to. **Bold** marks the higher value per row. † CQ5 remains below the moderate threshold (κ < 0.40) in both settings.

This procedure was applied uniformly across all ten annotation items prior to computing composite dimension means and overall scores. All analyses reported in Table 4 and Tables 10–11 derive from these majority-voted scores.

**Annotation dimensions and items.** All Likert scales ranged from 1 (*Strongly Disagree*) to 5 (*Strongly Agree*). Reverse-coded items were transformed as $6 - x$ before compositing. Cross-turn contradiction (CF3) was coded categorically (Yes / No / Cannot Determine) and excluded from the continuous composite. Epistemic Accuracy items (EA1-EA3) were collected but excluded from final composite scores following calibration-phase review, which identified systematic annotator disagreement on the plausibility judgements required; we do not report EA results.

**Content Quality (CQ).** Assesses whether the response is well-formed and task-fulfilling independent of its emotional character; this is a necessary baseline condition for any steered output to be usable in practice. Composite: CQ = mean(CQ3, 6−CQ5, CQ6).

- CQ3 Fulfills the user's request.
- CQ5 Contains internal contradictions. (reverse-coded)†
- CQ6 Avoids repeating the same idea unnecessarily.

**Contextual Fit (CF).** Assesses cross-turn coherence: whether steering disrupts the model's grounding in the preceding dialogue, which would undermine the multi-turn persistence claim. Composite: CF = mean(CF1, CF2).

- CF1 Logically follows from the previous turn.
- CF2 Maintains consistency with earlier information.
- CF3 Contradicts prior dialogue context. (categorical; excluded from composite)

**Emotional Quality (EQ).** Directly evaluates the target behavior of both emotional steering tasks: whether affective tone is appropriate, proportionate, and consistent rather than simply shifted in a detectable distributional direction. Composite: EQ = mean(EQ1, EQ2, EQ3).

- EQ1 Emotional tone matches the situation.
- EQ2 Emotional intensity is proportionate to the context.
- EQ3 Emotional tone remains consistent throughout the response.

**Perceived Naturalness (PN).** Assesses whether steered outputs retain human-like phrasing or exhibit the repetitive, stilted quality characteristic of over-steered generation. Composite: PN = mean(PN3, 6−PN4).

- PN3 Tone feels conversational.
- PN4 Contains unnecessary repetition. (reverse-coded)

**Overall score.** The overall mean is computed as Overall = mean(CQ, CF, EQ, PN).

**Statistical analysis.** Pairwise condition comparisons used independent-samples Welch's t-tests on composite dimension means (CQ, CF, EQ, PN, Overall), with Benjamini–Hochberg FDR correction applied separately within the single-turn tier (30 tests) and the multi-turn tier (50 tests). Within single-turn conditions, STAR (steered) is the only condition to produce significant differences after correction: it outperforms Prompt Baseline (S) on CQ, CF, PN, and Overall ($p_{adj} \leq .031$), and outperforms Global Steering on EQ and Overall ($p_{adj}$ = .031). No multi-turn pairwise comparison survives correction ($p_{adj} \geq .97$ throughout), nor does any cross-turn comparison in the fallback analysis. These results should be interpreted in the context of per-condition sample sizes of N = 10–46: the observed mean differences are substantively large relative to the scale range—up to $\Delta = +1.33$ overall between STAR (steered) and Prompt Baseline (S)—but the study is underpowered to detect moderate effects at these sample sizes, and results in the range $.01 < p_{adj} < .05$ should be treated as suggestive rather than confirmatory. Full pairwise statistics are reported in Tables 10 and 11.

**Inter-annotator agreement.** Full IAA results for the empathy annotation are reported in Table 7. ICC2k ranged from 0.74 (CQ5, the reverse-coded contradiction item) to 0.93 (CQ3, CQ6) across dimensions, with Cronbach's $\alpha \geq 0.93$ per annotator across all ten items. The negotiation annotation yielded ICC2k = 0.88-0.93 (Table 8). Agreement was substantially higher for single-turn than multi-turn items across every dimension (mean κ: single = 0.76, multi = 0.64; Table 9), consistent with single-turn responses providing an unambiguous context window for each judgment. Rolling kappa across the 200-item empathy session confirmed that agreement was stable throughout, with no evidence of fatigue-driven degradation for the nine main dimensions (Figure 9).

† CQ5 was the weakest dimension across all annotation phases (ICC2k: 0.63 at calibration, 0.74 on the full empathy round, 0.93 on negotiation) and showed high volatility in the rolling kappa analysis (Figure 9b). Scores on the CQ composite should therefore be interpreted with greater caution than scores on CF, EQ, or PN.

### F.3 Extended Negotiation Results

Tables 12,13, 14, and 15 report the full statistical breakdown of buyer behavior across bilateral, asymmetric, single-turn, and prompt-baseline comparisons on the CraigslistBargain corpus. Three findings are consistent across all four tables and warrant emphasis.

**Behavioral shifts are agent-local, not dyadic.** The clearest causal test of this claim is the null result for SS vs. US in Table 14: adding a steered first-turn agent on top of an already-steered system produces no further categorical change (all $V \leq 0.08$). By contrast, the same steered state produces large and significant shifts whenever it is compared against an unsteered baseline—UU vs. US and UU vs. SU both yield significance on five or six features. The implication is that STAR operates within each agent independently; the behavioral fingerprint of steering does not compound or suppress across turns. We cannot rule out that this null reflects a behavioural ceiling rather than strict agent-independence; distinguishing these interpretations would require conditions with weaker initial steering.

**Attribution-guided localization consistently outperforms global steering.** Across all four tables, local steering (LS / STAR) produces equal or larger effect sizes than global steering (GS) on every feature where the two are compared directly. The GS vs. LS comparison in Table 12 identifies hedging and apologizing as the features most sensitive to attribution-guided layer selection ($V$ = 0.11 and 0.12 respectively), consistent with the heatmap evidence that these behaviors are concentrated in a narrow band of mid-network layers. Global steering spreads the intervention uniformly and loses this resolution.

**Global steering converges with the prompt baseline; STAR does not.** Table 13 shows that GS and the prompt baseline (PB) are statistically indistinguishable on four of six categorical features ($V \leq 0.04$ on four items), whereas STAR (SS) retains significant separation from PB specifically on apologizing and indirect requests—the two markers most tightly linked to prosocial accommodation. This divergence is mechanistically informative: prompt instructions and unlocalized steering appear to achieve similar distributional shifts through similar representational pathways, while attribution-guided steering accesses a causally distinct locus.

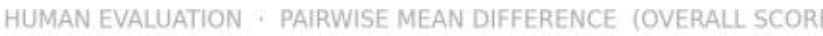

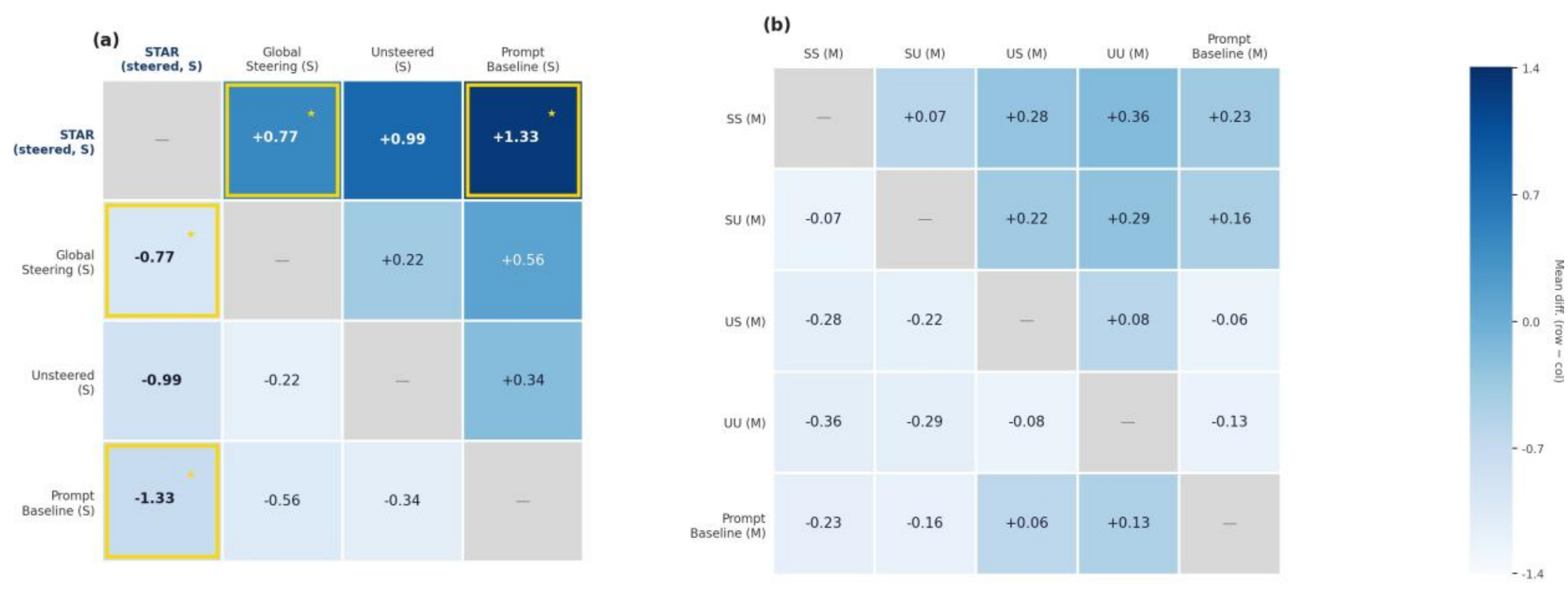

★ p_adj < 0.05 (Benjamini–Hochberg FDR correction, single-turn only)

Figure 10: Pairwise mean score differences across all evaluated conditions (N = 200 items, 1–5 Likert scale). Each cell reports the signed difference in overall mean score between the row and column condition (row − column); colour intensity encodes magnitude on a single blue ramp (light = near zero; dark = large magnitude), and diagonal cells are shown in grey. **(a)** Single-turn conditions only (STAR, Global Steering, Unsteered, Prompt Baseline). Gold-bordered cells marked with ⋆ denote comparisons significant after Benjamini–Hochberg FDR correction ($p_{adj} < 0.05$); STAR (steered) significantly outscores both Global Steering (+0.77) and Prompt Baseline (S) (+1.33). **(b)** Multi-turn conditions only (SS, SU, US, UU, Prompt Baseline); no pairwise comparison reaches significance ($p_{adj} \geq .97$ throughout).

| **Single-turn pairwise comparisons** | | *Welch's t-test; $p_{adj}$ BH-corrected across all 30 tests* | | | | | |
|---|---|---|---|---|---|---|---|
| **Comparison** | **Dim.** | $\bar{x}_A$ | $\bar{x}_B$ | $\Delta$ | $t$ | $p$ | $p_{adj}$ |
| *STAR (steered) vs. Global Steering* $n_A = 46,\ n_B = 17$ | | | | | | | |
| | CQ | 3.68 | 3.04 | +0.64 | 2.08 | .046 | .115 |
| | CF | 3.38 | 2.47 | +0.91 | 2.66 | .012 | .052 |
| | EQ | 3.50 | 2.61 | +0.89 | 3.03 | **.004** | **.031**$^{\bullet}$ |
| | PN | 3.50 | 2.85 | +0.65 | 1.74 | .092 | .185 |
| | Overall | 3.53 | 2.76 | +0.77 | 2.91 | **.006** | **.031**$^{\bullet}$ |
| *STAR (steered) vs. Unsteered* $n_A = 46,\ n_B = 13$ | | | | | | | |
| | CQ | 3.68 | 2.87 | +0.81 | 1.81 | .089 | .185 |
| | CF | 3.38 | 2.42 | +0.96 | 1.97 | .065 | .151 |
| | EQ | 3.50 | 2.38 | +1.12 | 2.31 | .034 | .114 |
| | PN | 3.50 | 2.38 | +1.12 | 2.23 | .040 | .115 |
| | Overall | 3.53 | 2.54 | +0.99 | 2.21 | .042 | .115 |
| *STAR (steered) vs. Prompt Baseline* $n_A = 46,\ n_B = 12$ | | | | | | | |
| | CQ | 3.68 | 2.36 | +1.32 | 3.76 | **.001** | **.024**$^{\bullet}$ |
| | CF | 3.38 | 2.12 | +1.26 | 3.60 | **.002** | **.024**$^{\bullet}$ |
| | EQ | 3.50 | 2.22 | +1.28 | 2.72 | .015 | .058 |
| | PN | 3.50 | 2.00 | +1.50 | 3.20 | **.006** | **.031**$^{\bullet}$ |
| | Overall | 3.53 | 2.20 | +1.33 | 3.55 | **.003** | **.026**$^{\bullet}$ |
| *Global Steering vs. Unsteered* $n_A = 17,\ n_B = 13$ | | | | | | | |
| | CQ | 3.04 | 2.87 | +0.17 | 0.34 | .735 | .788 |
| | CF | 2.47 | 2.42 | +0.05 | 0.09 | .929 | .929 |
| | EQ | 2.61 | 2.38 | +0.22 | 0.45 | .662 | .735 |
| | PN | 2.85 | 2.38 | +0.47 | 0.84 | .411 | .587 |
| | Overall | 2.76 | 2.54 | +0.22 | 0.47 | .645 | .735 |
| *Global Steering vs. Prompt Baseline* $n_A = 17,\ n_B = 12$ | | | | | | | |
| | CQ | 3.04 | 2.36 | +0.68 | 1.69 | .105 | .197 |
| | CF | 2.47 | 2.12 | +0.35 | 0.85 | .401 | .587 |
| | EQ | 2.61 | 2.22 | +0.39 | 0.79 | .440 | .600 |
| | PN | 2.85 | 2.00 | +0.85 | 1.60 | .123 | .217 |
| | Overall | 2.76 | 2.20 | +0.56 | 1.40 | .179 | .298 |
| *Unsteered vs. Prompt Baseline* $n_A = 13,\ n_B = 12$ | | | | | | | |
| | CQ | 2.87 | 2.36 | +0.51 | 0.99 | .334 | .527 |
| | CF | 2.42 | 2.12 | +0.30 | 0.56 | .581 | .697 |
| | EQ | 2.38 | 2.22 | +0.16 | 0.26 | .796 | .823 |
| | PN | 2.38 | 2.00 | +0.38 | 0.61 | .546 | .683 |
| | Overall | 2.54 | 2.20 | +0.34 | 0.63 | .537 | .683 |

*Dimensions:* CQ = Content Quality; CF = Contextual Fit; EQ = Emotional Quality; PN = Perceived Naturalness; Overall = mean across all ten items. $\Delta = \bar{x}_A - \bar{x}_B$; positive values favour group $A$. Highlighted rows denote $p_{adj} < 0.05$; $^{\bullet}$ marks the corrected $p$-value.

Table 10: Pairwise Welch's $t$-tests for single-turn conditions across five composite dimensions ($N_{total} = 88$; per-condition $n = 12$–$46$). STAR (steered) is the only condition to produce significant differences after Benjamini–Hochberg correction, outperforming Prompt Baseline on CQ, CF, PN, and Overall ($p_{adj} \leq .031$) and Global Steering on EQ and Overall ($p_{adj} = .031$). All remaining comparisons are non-significant ($p_{adj} \geq .052$).

| **Multi-turn pairwise comparisons** | *Welch's t-test; $p_{adj}$ BH-corrected across all 50 tests* | | | | | | |
|---|---|---|---|---|---|---|---|
| **Comparison** | **Dim.** | $\bar{x}_A$ | $\bar{x}_B$ | $\Delta$ | $t$ | $p$ | $p_{adj}$ |
| *SS vs. SU* $n_A = 38,\ n_B = 20$ | | | | | | | |
| | CQ | 3.58 | 3.22 | +0.36 | 0.93 | .357 | .974 |
| | CF | 3.33 | 3.38 | −0.05 | −0.11 | .911 | .974 |
| | EQ | 3.39 | 3.18 | +0.21 | 0.53 | .601 | .974 |
| | PN | 3.41 | 3.88 | −0.47 | −1.13 | .265 | .974 |
| | Overall | 3.44 | 3.37 | +0.07 | 0.18 | .855 | .974 |
| *SS vs. US* $n_A = 38,\ n_B = 20$ | | | | | | | |
| | CQ | 3.58 | 3.38 | +0.20 | 0.59 | .560 | .974 |
| | CF | 3.33 | 3.25 | +0.08 | 0.21 | .836 | .974 |
| | EQ | 3.39 | 3.05 | +0.34 | 0.88 | .382 | .974 |
| | PN | 3.41 | 2.88 | +0.53 | 1.33 | .191 | .974 |
| | Overall | 3.44 | 3.16 | +0.28 | 0.83 | .409 | .974 |
| *SS vs. UU* $n_A = 38,\ n_B = 24$ | | | | | | | |
| | CQ | 3.58 | 3.06 | +0.52 | 1.36 | .180 | .974 |
| | CF | 3.33 | 3.00 | +0.33 | 0.84 | .408 | .974 |
| | EQ | 3.39 | 3.07 | +0.33 | 0.80 | .427 | .974 |
| | PN | 3.41 | 3.19 | +0.22 | 0.50 | .616 | .974 |
| | Overall | 3.44 | 3.08 | +0.36 | 0.94 | .352 | .974 |
| *SS vs. Prompt Baseline* $n_A = 38,\ n_B = 10$ | | | | | | | |
| | CQ | 3.58 | 3.40 | +0.18 | 0.37 | .718 | .974 |
| | CF | 3.33 | 3.35 | −0.02 | −0.04 | .971 | .974 |
| | EQ | 3.39 | 3.13 | +0.26 | 0.45 | .661 | .974 |
| | PN | 3.41 | 2.90 | +0.51 | 0.85 | .412 | .974 |
| | Overall | 3.44 | 3.21 | +0.23 | 0.43 | .675 | .974 |
| *SU vs. US* $n_A = 20,\ n_B = 20$ | | | | | | | |
| | CQ | 3.22 | 3.38 | −0.17 | −0.40 | .693 | .974 |
| | CF | 3.38 | 3.25 | +0.13 | 0.28 | .784 | .974 |
| | EQ | 3.18 | 3.05 | +0.13 | 0.30 | .764 | .974 |
| | PN | 3.88 | 2.88 | +1.00 | 2.24 | .031 | .974 |
| | Overall | 3.37 | 3.16 | +0.21 | 0.53 | .602 | .974 |
| *SU vs. UU* $n_A = 20,\ n_B = 24$ | | | | | | | |
| | CQ | 3.22 | 3.06 | +0.16 | 0.35 | .728 | .974 |
| | CF | 3.38 | 3.00 | +0.38 | 0.80 | .426 | .974 |
| | EQ | 3.18 | 3.07 | +0.11 | 0.25 | .804 | .974 |
| | PN | 3.88 | 3.19 | +0.69 | 1.43 | .159 | .974 |
| | Overall | 3.37 | 3.08 | +0.29 | 0.66 | .514 | .974 |
| *SU vs. Prompt Baseline* $n_A = 20,\ n_B = 10$ | | | | | | | |
| | CQ | 3.22 | 3.40 | −0.18 | −0.33 | .741 | .974 |
| | CF | 3.38 | 3.35 | +0.03 | 0.04 | .969 | .974 |
| | EQ | 3.18 | 3.13 | +0.05 | 0.08 | .937 | .974 |
| | PN | 3.88 | 2.90 | +0.98 | 1.54 | .142 | .974 |
| | Overall | 3.37 | 3.21 | +0.16 | 0.28 | .786 | .974 |
| *US vs. UU* $n_A = 20,\ n_B = 24$ | | | | | | | |
| | CQ | 3.38 | 3.06 | +0.33 | 0.79 | .435 | .974 |
| | CF | 3.25 | 3.00 | +0.25 | 0.57 | .571 | .974 |
| | EQ | 3.05 | 3.07 | −0.02 | −0.04 | .965 | .974 |
| | PN | 2.88 | 3.19 | −0.31 | −0.67 | .508 | .974 |
| | Overall | 3.16 | 3.08 | +0.08 | 0.19 | .849 | .974 |
| *US vs. Prompt Baseline* $n_A = 20,\ n_B = 10$ | | | | | | | |
| | CQ | 3.38 | 3.40 | −0.02 | −0.03 | .974 | .974 |
| | CF | 3.25 | 3.35 | −0.10 | −0.16 | .871 | .974 |
| | EQ | 3.05 | 3.13 | −0.08 | −0.14 | .893 | .974 |
| | PN | 2.88 | 2.90 | −0.03 | −0.04 | .969 | .974 |
| | Overall | 3.16 | 3.21 | −0.05 | −0.10 | .923 | .974 |
| *UU vs. Prompt Baseline* $n_A = 24,\ n_B = 10$ | | | | | | | |
| | CQ | 3.06 | 3.40 | −0.34 | −0.63 | .535 | .974 |
| | CF | 3.00 | 3.35 | −0.35 | −0.57 | .578 | .974 |
| | EQ | 3.07 | 3.13 | −0.06 | −0.10 | .919 | .974 |
| | PN | 3.19 | 2.90 | +0.29 | 0.44 | .662 | .974 |
| | Overall | 3.08 | 3.21 | −0.13 | −0.23 | .821 | .974 |

*Dimensions:* CQ = Content Quality; CF = Contextual Fit; EQ = Emotional Quality; PN = Perceived Naturalness; Overall = mean across all ten items. SS/SU/US/UU denote the steered (S) or unsteered (U) status of the first and second generation turns respectively. $\Delta = \bar{x}_A - \bar{x}_B$; positive values favour group $A$. No rows are highlighted as no comparison reaches $p_{adj} < 0.05$.

Table 11: Pairwise Welch's $t$-tests for multi-turn conditions across five composite dimensions ($N_{total} = 112$; per-condition $n$ = 10–38). No comparison reaches significance after Benjamini–Hochberg correction ($p_{adj} \geq .97$ throughout), indicating that the four multi-turn steering configurations and the multi-turn prompt baseline are statistically indistinguishable on all evaluated dimensions. Note that SU vs. US yields a nominally significant raw $p = .031$ on PN, which does not survive correction.

**Comparison 1:** Unsteered vs. Local Steered (LS / STAR)

| Feature | Unst. | LS | p | Effect |
|---|---|---|---|---|
| *Categorical (%); BH-corrected p; Effect = Cramér's V (size)* | | | | |
| Gratitude | 15 | 44 | **<.001** | **0.31 (Med.)** |
| Hedges | 22 | 41 | **<.001** | **0.20 (Sm.)** |
| Apologizing | 8 | 31 | **<.001** | **0.28 (Sm.)** |
| Indirect Requests | 11 | 37 | **<.001** | **0.30 (Med.)** |
| Directness | 48 | 19 | **<.001** | **0.30 (Med.)** |
| Dismissiveness | 23 | 9 | **<.001** | **0.19 (Sm.)** |
| *Continuous; Effect = BH-corrected q (Welch's t-test)* | | | | |
| Agreement Rate | 3.50 | 4.20 | 0.668 | 0.933 |
| Price Improv.‡ | −0.31 | 0.52 | **0.006** | **0.021** |

**Comparison 2:** Unsteered vs. Global Steered (GS)

| Feature | Unst. | GS | p | Effect |
|---|---|---|---|---|
| *Categorical (%); BH-corrected p; Effect = Cramér's V (size)* | | | | |
| Gratitude | 15 | 41 | **<.001** | **0.27 (Sm.)** |
| Hedges | 22 | 31 | **0.025** | **0.10 (Sm.)** |
| Apologizing | 8 | 20 | **0.002** | **0.13 (Sm.)** |
| Indirect Requests | 11 | 33 | **<.001** | **0.24 (Sm.)** |
| Directness | 48 | 22 | **<.001** | **0.26 (Sm.)** |
| Dismissiveness | 23 | 11 | **0.005** | **0.12 (Sm.)** |
| *Continuous; Effect = BH-corrected q (Welch's t-test)* | | | | |
| Agreement Rate | 3.50 | 4.05 | 0.550 | 0.850 |
| Price Improv.‡ | −0.31 | 0.33 | **0.015** | **0.050** |

**Comparison 3:** Global Steered (GS) vs. Local Steered (LS)

| Feature | GS | LS | p | Effect |
|---|---|---|---|---|
| *Categorical (%); BH-corrected p; Effect = Cramér's V (size)* | | | | |
| Gratitude | 41 | 44 | 0.350 | 0.04 (Tr.) |
| Hedges | 31 | 41 | **0.015** | **0.11 (Sm.)** |
| Apologizing | 20 | 31 | **0.008** | **0.12 (Sm.)** |
| Indirect Requests | 33 | 37 | 0.210 | 0.06 (Tr.) |
| Directness | 22 | 19 | 0.280 | 0.05 (Tr.) |
| Dismissiveness | 11 | 9 | 0.450 | 0.03 (Tr.) |
| *Continuous; Effect = BH-corrected q (Welch's t-test)* | | | | |
| Agreement Rate | 4.05 | 4.20 | 0.700 | 0.900 |
| Price Improv.‡ | 0.33 | 0.52 | 0.250 | 0.450 |

*Abbreviations:* LS = STAR (local, attribution-guided steering); GS = global (unlocalized) steering; Unst. = unsteered. Sm. = Small; Med. = Medium; Tr. = Trivial. † Diagnostic proxy for verbosity; not treated as a primary outcome measure. ‡ Positive values indicate buyer price below dataset baseline (buyer gain); negative values indicate buyer pays above baseline.

Table 12: Local steering (STAR) shifts buyer behavior toward accommodation across all six politeness markers in single-turn negotiations, with medium effect sizes on gratitude ($V = 0.31$), indirect requests ($V = 0.30$), and directness ($V = 0.30$). Global steering (GS) produces a broadly similar shift but with uniformly small effect sizes; the GS vs. LS comparison reveals that attribution-guided localization specifically advantages hedging and apologizing—the markers most sensitive to layer-level injection. Among continuous metrics, only price improvement reaches significance under both LS and GS.

| **Multi-Turn Comparison 1:** Steered–Steered (SS / STAR) vs. Prompt Baseline (PB) | | | | |
|---|---|---|---|---|
| **Feature** | **SS** | **PB** | **p** | **Effect** |
| *Categorical (%); BH-corrected p; Effect = Cramér's V (size)* | | | | |
| Gratitude | 51 | 38 | **0.003** | **0.13 (Sm.)** |
| Hedges | 48 | 36 | **0.005** | **0.12 (Sm.)** |
| Apologizing | 36 | 20 | **<.001** | **0.18 (Sm.)** |
| Indirect Requests | 45 | 27 | **<.001** | **0.18 (Sm.)** |
| Directness | 21 | 30 | **0.018** | **0.10 (Sm.)** |
| Dismissiveness | 8 | 14 | **0.040** | 0.09 (Tr.) |
| *Continuous; Effect = BH-corrected q (Welch's t-test)* | | | | |
| Agreement Rate | 6.64 | 4.08 | 0.207 | 0.496 |
| Price Improv.[‡] | 0.71 | 0.40 | 0.471 | 0.789 |

| **Multi-Turn Comparison 2:** Unsteered–Unsteered (UU) vs. Prompt Baseline (PB) | | | | |
|---|---|---|---|---|
| **Feature** | **UU** | **PB** | **p** | **Effect** |
| *Continuous; Effect = BH-corrected q (Welch's t-test)* | | | | |
| Agreement Rate | 4.80 | 4.08 | 0.700 | 0.933 |
| Price Improv.[‡] | 0.25 | 0.40 | 0.564 | 0.850 |

| **Single-Turn Comparison 1:** Local Steered (LS / STAR) vs. Prompt Baseline (PB) | | | | |
|---|---|---|---|---|
| **Feature** | **LS** | **PB** | **p** | **Effect** |
| *Categorical (%); BH-corrected p; Effect = Cramér's V (size)* | | | | |
| Gratitude | 44 | 38 | 0.163 | 0.06 (Tr.) |
| Hedges | 41 | 33 | 0.062 | 0.08 (Tr.) |
| Apologizing | 31 | 18 | **<.001** | **0.15 (Sm.)** |
| Indirect Requests | 37 | 24 | **0.001** | **0.14 (Sm.)** |
| Directness | 19 | 30 | **0.004** | **0.12 (Sm.)** |
| Dismissiveness | 9 | 12 | 0.263 | 0.05 (Tr.) |
| *Continuous; Effect = BH-corrected q (Welch's t-test)* | | | | |
| Agreement Rate | 4.20 | 4.08 | 0.947 | 1.000 |
| Price Improv.[‡] | 0.52 | 0.40 | 0.726 | 0.933 |

| **Single-Turn Comparison 2:** Global Steered (GS) vs. Prompt Baseline (PB) | | | | |
|---|---|---|---|---|
| **Feature** | **GS** | **PB** | **p** | **Effect** |
| *Categorical (%); BH-corrected p; Effect = Cramér's V (size)* | | | | |
| Gratitude | 41 | 38 | 0.380 | 0.04 (Tr.) |
| Hedges | 31 | 33 | 0.520 | 0.03 (Tr.) |
| Apologizing | 20 | 18 | 0.460 | 0.03 (Tr.) |
| Indirect Requests | 33 | 24 | **0.030** | **0.10 (Sm.)** |
| Directness | 22 | 30 | **0.045** | 0.09 (Tr.) |
| Dismissiveness | 11 | 12 | 0.750 | 0.02 (Tr.) |
| *Continuous; Effect = BH-corrected q (Welch's t-test)* | | | | |
| Agreement Rate | 4.05 | 4.08 | 0.950 | 1.000 |
| Price Improv.[‡] | 0.33 | 0.40 | 0.810 | 0.950 |

*Abbreviations:* LS = STAR (local, attribution-guided steering); GS = global (unlocalized) steering; UU = both agents unsteered; SS = both agents steered (STAR); PB = prompt (zero-shot instruction) baseline. Sm. = Small; Med. = Medium; Tr. = Trivial. [†] Diagnostic proxy for verbosity; not treated as a primary outcome measure. [‡] Positive values indicate buyer price below dataset baseline (buyer gain); negative values indicate buyer pays above baseline.

Table 13: Activation steering and prompt engineering achieve behavioral change through qualitatively different mechanisms. Global steering (GS) converges with the prompt baseline on most categorical features—no significant difference survives on four of six markers—suggesting that unlocalized steering replicates prompt-level distributional effects. Local steering (STAR) differentiates from PB specifically on apologizing and indirect requests, where attribution patching targets causally active layers. Economic outcomes (agreement rate, price improvement) do not differ significantly across any baseline comparison.

| **Comparison 1:** Unsteered–Unsteered (UU) vs. Unsteered–Steered (US) | | | | |
|---|---|---|---|---|
| **Feature** | **UU** | **US** | **p** | **Effect** |
| *Categorical (%); BH-corrected p; Effect = Cramér's V (size)* | | | | |
| Gratitude | 18 | 44 | **<.001** | **0.28 (Sm.)** |
| Hedges | 25 | 43 | **<.001** | **0.19 (Sm.)** |
| Apologizing | 10 | 28 | **<.001** | **0.23 (Sm.)** |
| Indirect Requests | 12 | 39 | **<.001** | **0.30 (Med.)** |
| Directness | 47 | 26 | **<.001** | **0.21 (Sm.)** |
| Dismissiveness | 22 | 12 | **0.003** | **0.13 (Sm.)** |
| *Continuous; Effect = BH-corrected q (Welch's t-test)* | | | | |
| Agreement Rate | 4.80 | 4.80 | 1.000 | 1.000 |
| Price Improv.[‡] | 0.25 | 0.38 | 0.694 | 0.933 |
| **Comparison 2:** Unsteered–Unsteered (UU) vs. Steered–Unsteered (SU) | | | | |
| **Feature** | **UU** | **SU** | **p** | **Effect** |
| *Categorical (%); BH-corrected p; Effect = Cramér's V (size)* | | | | |
| Gratitude | 18 | 39 | **<.001** | **0.23 (Sm.)** |
| Hedges | 25 | 34 | 0.030 | 0.09 (Tr.) |
| Apologizing | 10 | 19 | **0.003** | **0.13 (Sm.)** |
| Indirect Requests | 12 | 26 | **<.001** | **0.17 (Sm.)** |
| Directness | 47 | 35 | **0.005** | **0.12 (Sm.)** |
| Dismissiveness | 22 | 17 | 0.159 | 0.06 (Tr.) |
| **Null Results:** SS vs. US and PB vs. SU *(categorical features; all effects trivial)* | | | | |
| **Feature** | **SS** | **US** | **p** | **Effect** |
| *SS vs. US: no feature reaches significance; all Cramér's* $V \leq 0.08$ | | | | |
| Gratitude | 51 | 44 | 0.122 | 0.07 (Tr.) |
| Hedges | 48 | 43 | 0.263 | 0.05 (Tr.) |
| Apologizing | 36 | 28 | 0.054 | 0.08 (Tr.) |
| Indirect Requests | 45 | 39 | 0.192 | 0.06 (Tr.) |
| Directness | 21 | 26 | 0.189 | 0.06 (Tr.) |
| Dismissiveness | 8 | 12 | 0.155 | 0.06 (Tr.) |
| **Feature** | **PB** | **SU** | **p** | **Effect** |
| *PB vs. SU: no feature reaches significance; all Cramér's* $V \leq 0.05$ | | | | |
| Gratitude | 38 | 39 | 0.860 | 0.01 (Tr.) |
| Hedges | 36 | 34 | 0.653 | 0.02 (Tr.) |
| Apologizing | 20 | 19 | 0.914 | 0.00 (Tr.) |
| Indirect Requests | 27 | 26 | 0.923 | 0.00 (Tr.) |
| Directness | 30 | 35 | 0.272 | 0.05 (Tr.) |
| Dismissiveness | 14 | 17 | 0.406 | 0.04 (Tr.) |
| **Comparison 3:** Prompt Baseline (PB) vs. Unsteered–Steered (US) | | | | |
| **Feature** | **PB** | **US** | **p** | **Effect** |
| *Continuous; Effect = BH-corrected q (Welch's t-test)* | | | | |
| Agreement Rate | 4.08 | 4.80 | 0.700 | 0.933 |
| Price Improv.[‡] | 0.40 | 0.38 | 0.953 | 1.000 |

*Abbreviations:* UU = both unsteered; SS = both steered (STAR); US = unsteered initiator, steered responder; SU = steered initiator, unsteered responder; PB = prompt (zero-shot instruction) baseline. Sm. = Small; Med. = Medium; Tr. = Trivial. [†] Diagnostic proxy for verbosity; not treated as a primary outcome measure. [‡] Positive values indicate buyer price below dataset baseline (buyer gain).

Table 14: Asymmetric steering conditions show that behavioral shifts are driven by each agent's own steering state. Both UU vs. US and UU vs. SU produce significant categorical shifts, confirming that injection takes effect regardless of turn position. The null results for SS vs. US (all $V \leq 0.08$) and PB vs. SU (all $V \leq 0.05$) confirm that the responder's steered state does not independently modulate behavior: adding steering to the second turn does not differentiate SS from US, and conditioning on the prompt baseline does not differentiate PB from SU. The PB vs. US continuous comparison replicates the structural profile seen wherever the prompt baseline is involved—lower question rate and shorter turns relative to any activation-steered condition.

| Comparison: Unsteered–Unsteered (UU) vs. Steered–Steered (SS / STAR) | | | | |
|---|---|---|---|---|
| **Feature** | **UU** | **SS** | **p** | **Effect** |
| *Categorical (%); BH-corrected p; Effect = Cramér's V (size)* | | | | |
| Gratitude | 18 | 51 | **<.001** | **0.34 (Med.)** |
| Hedges | 25 | 48 | **<.001** | **0.24 (Sm.)** |
| Apologizing | 10 | 36 | **<.001** | **0.31 (Med.)** |
| Indirect Requests | 12 | 45 | **<.001** | **0.36 (Med.)** |
| Directness | 47 | 21 | **<.001** | **0.27 (Sm.)** |
| Dismissiveness | 22 | 8 | **<.001** | **0.19 (Sm.)** |
| *Continuous; Effect = BH-corrected q (Welch's t-test)* | | | | |
| Agreement Rate | 4.80 | 6.64 | 0.356 | 0.694 |
| Price Improv.‡ | 0.25 | 0.71 | 0.280 | 0.593 |

*Abbreviations:* UU = both agents unsteered; SS = both agents steered (STAR). Sm. = Small; Med. = Medium. † Diagnostic proxy for verbosity; not treated as a primary outcome measure. ‡ Positive values indicate buyer price below dataset baseline (buyer gain).

Table 15: Bilateral activation steering (SS) sustains consistent shifts in buyer accommodation behavior across all six categorical features in multi-turn negotiations, with medium effect sizes on gratitude ($V = 0.34$), apologizing ($V = 0.31$), and indirect requests ($V = 0.36$). Continuous outcome metrics are uniformly non-significant, indicating that STAR reshapes *how* the buyer negotiates—shifting toward cooperative politeness strategies—without altering *whether* agreements are reached or price outcomes.